\documentclass[fleqn]{article}

\usepackage{iclr2020_conference,times}
\iclrfinalcopy % NB this is for arXiv.
\usepackage{microtype}
\usepackage{color}
\usepackage{mathtools}
\usepackage{amsfonts}
\usepackage{amssymb}
\usepackage{graphicx}
\graphicspath{{./imgs/}}
\usepackage{subcaption}
\usepackage{algorithm}
\usepackage[noend]{algpseudocode}
\usepackage{tikz} % for tables
\usepackage{collcell} % for tables
\usepackage{xstring} % for tables
\usepackage{booktabs} % for tables
\usepackage{multirow}
\usepackage{hyperref}
\usepackage{subcaption}
\usepackage[utf8]{inputenc} % allow utf-8 input
\usepackage{nicefrac}       % compact symbols for 1/2, etc.
\usepackage[T1]{fontenc}
\usepackage{float}
\usepackage{wrapfig}
\usepackage{natbib}
\usepackage{tabularx}

\usepackage[subtle]{savetrees} % encourage to squash things onto page

\newcolumntype{Y}{>{\centering\arraybackslash\tiny}X}

\hypersetup{
    pdfauthor={Isaac Dunn, Hadrien Pouget, Tom Melham, Daniel Kroening},   % authors
    pdftitle={Adaptive Generation of Unrestricted Adversarial Inputs},   % document title
    colorlinks=true,                          % links are coloured, not boxed
    linkcolor=black,                          % colour of internal links
    citecolor=blue,                           % colour of links to bibliography
    urlcolor=blue                             % colour of external links
}

\title{Adaptive Generation of Unrestricted \\Adversarial Inputs}

\author{%
	Isaac Dunn, Hadrien Pouget,
	 Tom Melham, Daniel Kroening \\
	Department of Computer Science,
	University of Oxford\\
	\texttt{Isaac.Dunn@cs.ox.ac.uk}
}

\DeclareMathOperator*{\argmax}{argmax}

\newcommand{\ColourCellBox}[4]{% cell label, colour percentage, first colour, second colour
		\hspace{-0.33em}\colorbox{#3!#2!#4}{#1}%
}
\newcommand{\ApplyAbtestGradient}[1]{% takes cell label and draws cell, including colour
	\newcommand*{\MinNumber}{0}% lower bound on input
	\newcommand*{\MaxNumber}{54}% upper bound on input
	\IfInteger{#1}{% if an integer, coloured-in cell
		\pgfmathdectobase\firstarg{#1}{10}
		\pgfmathsetmacro{\PercentColour}{100.0*(\firstarg-\MinNumber)/(\MaxNumber-\MinNumber)}% calculate colour split proportion
		\ColourCellBox{#1}{\PercentColour}{purple}{white}
	}{\IfStrEq{#1}{XX}
		{\hspace{-0.33em}\colorbox{black}{00}}% plain black if X
		{\ColourCellBox{#1}{1}{white}{white}}% else plain white
	}
}
\newcommand{\ApplyOddoneoutGradient}[1]{% takes cell label and draws cell, including colour
	\newcommand*{\MinNumber}{00}%
	\newcommand*{\MaxNumber}{90}%
	\IfInteger{#1}{% if an integer, coloured-in cell
		\pgfmathdectobase\firstarg{#1}{10}
		\pgfmathsetmacro{\PercentColour}{100.0*(\firstarg-\MinNumber)/(\MaxNumber-\MinNumber)}%
		\ColourCellBox{#1}{\PercentColour}{green}{white}
	}{\IfStrEq{#1}{XX}
		{\hspace{-0.33em}\colorbox{black}{00}}% plain black if X
		{\ColourCellBox{#1}{1}{white}{white}}% else plain white
	}
}
\newcommand{\ApplyLabelmeGradient}[1]{% takes cell label and draws cell, including colour
	\newcommand*{\MinNumber}{0}%
	\newcommand*{\MaxNumber}{100}%
	\IfInteger{#1}{% if an integer, coloured-in cell
		\pgfmathdectobase\firstarg{#1}{10}
		\pgfmathsetmacro{\PercentColour}{(100.0-100.0*(\firstarg-\MinNumber)/(\MaxNumber-\MinNumber))^(0.5) * (100.0^(1-0.5))}%
		\ColourCellBox{#1}{\PercentColour}{red}{white}
	}{\IfStrEq{#1}{XX}
		{\hspace{-0.33em}\colorbox{black}{00}}% plain black if X
		{\ColourCellBox{#1}{1}{white}{white}}% else plain white
	}
}
\newcommand{\ApplyTransferGradient}[1]{% takes cell label and draws cell, including colour
	\newcommand*{\MinNumber}{0}%
	\newcommand*{\MaxNumber}{80}%
	\IfInteger{#1}{% if an integer, coloured-in cell
		\pgfmathdectobase\firstarg{#1}{10}
		\pgfmathsetmacro{\PercentColour}{(100.0-100.0*(\firstarg-\MinNumber)/(\MaxNumber-\MinNumber))^(0.5) * (100.0^(1-0.5))}%
		\ColourCellBox{#1}{\PercentColour}{white}{purple}
	}{\IfStrEq{#1}{XX}
		{\hspace{-0.33em}\colorbox{black}{00}}% plain black if X
		{\ColourCellBox{#1}{1}{white}{white}}% else plain white
	}
}

\newcolumntype{A}{>{\collectcell\ApplyAbtestGradient}c<{\endcollectcell}}
\newcolumntype{O}{>{\collectcell\ApplyOddoneoutGradient}c<{\endcollectcell}}
\newcolumntype{L}{>{\collectcell\ApplyLabelmeGradient}c<{\endcollectcell}}
\newcolumntype{T}{>{\collectcell\ApplyTransferGradient}c<{\endcollectcell}}
\renewcommand{\arraystretch}{0.9} % whitespace between rows (multiplicative factor)
\begin{document}

\maketitle

\begin{abstract}
Neural networks are vulnerable to
adversarially-constructed perturbations of their inputs.
Most research so far has considered perturbations
of a fixed magnitude under some $l_p$ norm.
Although studying these attacks is valuable,
there has been increasing interest in
the construction of---and robustness
to---\textit{unrestricted} attacks,
which are not constrained to a small and rather artificial
subset of all possible adversarial inputs.
We introduce a novel algorithm for generating
such unrestricted adversarial inputs which,
unlike prior work,
is \emph{adaptive}: it is able to tune
its attacks to the classifier being targeted.
It also offers a 400--2,000$\times$ speedup over the
existing state of the art.
We demonstrate our approach by generating unrestricted adversarial
inputs that fool classifiers robust to
perturbation-based attacks.
We also show that,
by virtue of being adaptive and unrestricted,
our attack is able to defeat
adversarial training against it.

\end{abstract}

\section{Introduction}
\setlength{\intextsep}{0pt}% vertical padding

Despite their dramatic successes in other respects,
neural networks are well-known to not be adversarially
robust.
\cite{DBLP:journals/corr/SzegedyZSBEGF13}
discovered that neural networks
are vulnerable to what they termed \textit{adversarial inputs}:
by adding carefully-chosen perturbations to
correctly-classified inputs, the accuracy
of any neural network could be
almost arbitrarily decreased.
Since then, the machine learning community has
rightly
focused a great deal of research effort on
this phenomenon.
Many early efforts to train more robust models
initially appeared promising, but 
have since been shown to be vulnerable to
new algorithms for constructing
adversarial perturbations
\citep{DBLP:journals/corr/abs-1909-08072}.
As a result, more attention has
been given to methods that
provide formal guarantees about
performance in the presence of
adversarial perturbations
\citep{DBLP:journals/corr/abs-1903-06758},
with the state of the art now
providing non-trivial guarantees
for the MNIST test set
\citep{
	DBLP:conf/icml/WongK18,
	DBLP:journals/corr/abs-1810-07481,
	DBLP:journals/corr/abs-1811-02625}.

However, almost all of this work has
focused exclusively on adversarial
perturbations whose magnitude is constrained
by an $l_p$ norm.
There is a growing acknowledgement that
this threat model is somewhat contrived:
such examples are
not a realistic security concern and also
occupy a vanishingly small fraction
of the set of potential adversarial inputs.
Therefore, there is a burgeoning interest
in adversarial attacks
that are \textit{unrestricted},
in the sense that they do not necessarily derive from
a perturbation of a natural
input~\citep{DBLP:journals/corr/abs-1809-08352, DBLP:conf/nips/SongSKE18}.

The main contribution of this paper is
a novel and general method to generate unrestricted
adversarial inputs.
In short, the training procedure for generative
adversarial networks (GANs) is modified
so that the generator network is rewarded
for producing data that are both realistic and
deceive a fixed target
network.
Our approach has four advantages over prior work:
\begin{enumerate}
	\item Our method is \emph{adaptive}
	in that it adjusts itself to best
	attack the specific network being targeted.
	For instance, adversarial training is ineffective
	against our approach.
	\item Our method is efficient (offering a 400--2000$\times$ speedup over prior work).
	\item Our method can easily be applied to \emph{any}
	existing GAN codebase and checkpoints,
	regardless of architecture, training procedure,
	or application domain.
	\item Our method therefore demonstrably scales to ImageNet.
\end{enumerate}

\section{Background: Generative Adversarial Networks}

Generative adversarial networks (GANs)
\citep{DBLP:conf/nips/GoodfellowPMXWOCB14}
are a class
of generative machine learning models
involving the simultaneous training of two neural
networks: a {generator}~$g$ and
a {discriminator} $d$.
Specifically, given
a dataset $D$ of samples drawn from
a probability distribution $p_D$,
the generator $g$
learns to transform
random noise $z$
drawn from a simple distribution
$p_z$
into an approximation of $p_D$.
The discriminator
network $d$ learns to
predict whether
a given example $x$
is drawn from the data distribution $p_D$
or was generated by $g$.
The generator and the discriminator are \emph{adversarial}
because they train simultaneously, with each being rewarded
for out-performing the other.

GANs'
training behaviours are notoriously temperamental,
and many
modifications to the original algorithm
have been proposed
\citep{DBLP:journals/corr/Goodfellow17}.
The Wasserstein GAN variant
\citep{DBLP:conf/icml/ArjovskyCB17}
aims to provide a more reliable gradient 
by designing the discriminator
(renamed `critic') to approximate the Wasserstein
distance between the distribution generated by $g_\theta$
and the data distribution $p_D$.
An additional
`gradient penalty' loss term $L_{\mathit{gp}}$
can be added to implement the constraint 
that the function 
be 1-Lipschitz continuous
\citep{DBLP:conf/nips/GulrajaniAADC17}.
The loss functions for this Wasserstein GAN with
gradient penalty (WGAN-GP) are:
$L_g = \mathop{\mathbb{E}}_{z \sim p_z}[-d(g(z))]$
and
$L_d = -L_g + \mathop{\mathbb{E}}_{x \sim p_D}[-d(x)]
	+\lambda L_{\mathit{gp}}$;
where the gradient penalty
$L_{\mathit{gp}} = \mathop{\mathbb{E}}_{\tilde{x} \sim p_I}
[(\Vert\nabla_{\tilde{x}}d_\phi(\tilde{x})\Vert_2 - 1)^2]$,
where $p_I$ denotes the distribution sampling
uniformly from the linear interpolations
between generated samples and examples from $p_D$.

\label{cgan}
The original proposal for a conditional generative
adversarial network (CGAN)
learns to generate samples from a conditional
distribution~\citep{DBLP:journals/corr/MirzaO14}
by simply passing the intended label $y$ for the
generated image to both the generator and
the discriminator.
An extension of this approach is the
auxiliary classifier generative
adversarial network (ACGAN)
\citep{DBLP:conf/icml/OdenaOS17},
in which the discriminator is modified
to also predict the label $y$ for the
input data. Both the generators are
trained to maximise the log-likelihood
of the correct label in addition to
optimising their usual objective.

\begin{figure*}[b]
	\centering
	\def\svgwidth{\textwidth}
	{ \large
		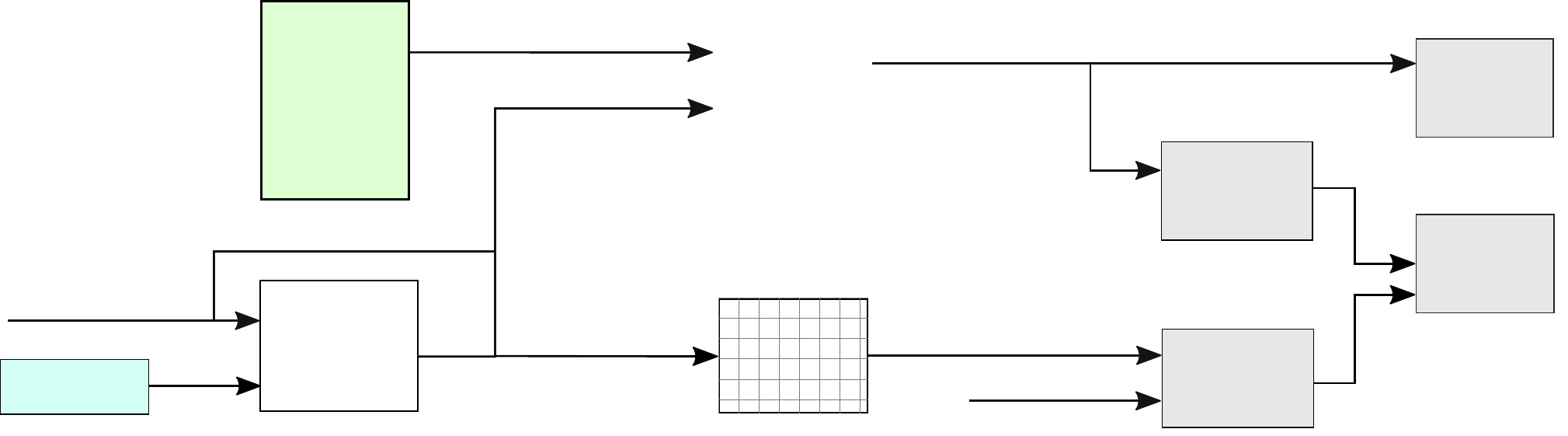 }
	\caption{Diagram showing main data paths in the
		forward computation of loss functions.
		During adversarial finetuning, the
		generator and discriminator are trained
		by backpropagating the gradients from
		$l_d$ and $l_\textit{finetune}$
		respectively.
		The target classifier, $f$,
		remains fixed.
}
	\label{main-architecture-diagram}
\end{figure*}

\section{Generating Unrestricted Adversarial Inputs}
\label{methods}

Suppose we have a trained
target classifier network $f\colon X \to \mathbb{R}^{|Y|}$
that attempts to approximate an oracle function $o\colon O \to Y$
(where $O \subseteq X$ is the oracle's domain)
by outputting a confidence $f(x)_c \in \mathbb{R}$ for each class
$c \in Y$.
As \cite{DBLP:conf/nips/SongSKE18} do,
we define an unrestricted adversarial example
to be any input $x \in O$ such that
the classifier's prediction is incorrect:
$\argmax_{c} f(x)_c	\neq o(x)$.
Unlike \citeauthor{DBLP:conf/nips/SongSKE18},
we consider the domain of the oracle to
be any input with a recognisable class,
not just realistic inputs.
This means we would still consider an unrealistic but
recognisable image an unrestricted adversarial example.
Nevertheless, we do carefully evaluate
how realistic our results are in
Section~\ref{realistic-experiments}.

Unrestricted adversarial examples are
a superset of conventional perturbation-based
adversarial examples (which are restricted to lie within a fixed distance
of some correctly-classified input from a test dataset).
While providing a vastly larger space of candidates,
a difficulty arises in determining that
the classification is incorrect;
we can no longer rely on the oracle-provided labels
from the test dataset.
We leverage generative models to solve this problem.

\subsection{Our Procedure}
\label{our-method}

We train a class-conditional GAN to generate 
unrestricted adversarial inputs.
This is achieved by simultaneously
minimising an ordinary GAN loss
and a new loss term. For an untargeted attack this term
rewards the generator if the examples
it generates are misclassified by
the target network:
$l_\textit{untargeted} = f(\hat{x})_y - \max_{c \neq y} f(\hat{x})_c$.
For a targeted attack, this term
rewards the generator
if the generated examples
are assigned the desired target label, $t$, by
the target classifier:
${l_\textit{targeted}=\max_{c \neq t} f(\hat{x})_c - f(\hat{x})_t}$.
Note that besides improving the quality of the
generated data, our use of a conditional GAN
and optimising for its ordinary loss function
allows these new loss terms to be
computed---otherwise, there
is no way of determining the label, $y$, for
each generated image.
This assumes that the true labels
of the generated data
match the intended labels passed as inputs
to the generator, an assumption
empirically validated in
Section~\ref{are-examples-adversarial}.

\subsection{Challenge: Conflicting Gradients}

Na\"{i}vely
optimising the sum of the two loss
terms cripples training.
There is no guarantee
that the loss landscape will
allow gradient-descent algorithms
to find optimum where the images
are both sufficiently realistic and 
adversarial,
and unfortunately
note that making an image 
adversarial seems likely to make
the image \textit{less} realistic,
not more.
This gives some intuition that the gradient
from $l_{\textit{ordinary}}$
may be pointing in a different direction
to the gradient from $l_\textit{(un)targeted}$.

A simple experiment suffices to
verify this intuition.
At each training step, the gradient vectors
from both loss terms were normalised,
then projected one onto the other.
That is, the scalar quantity
$
\frac{
	{\nabla l_\textit{ordinary}}
	\cdot
	{\nabla l_\textit{(un)targeted}}
}{
	\|\nabla l_\textit{ordinary}\|\|\nabla l_\textit{(un)targeted}\|
}
$
was computed.
Figure~\ref{local-conflict-from-start}
shows that this projection tends towards $-1$;
for reference, if the gradient vectors were
selected uniformly at random, the
magnitude of this projection would
rarely exceed 0.001.
In other words, as training progresses,
the gradients from these terms
tend towards pointing in
\emph{opposite} directions.
This makes joint optimisation
using gradient descent
a challenge.

\subsection{Strategies To Overcome Training Challenges}
\label{overcome-training-challenges}

\begin{figure}[b]
	\centering
	\begin{subfigure}[b]{0.49\linewidth}
		\centering
		\includegraphics[width=1\linewidth]{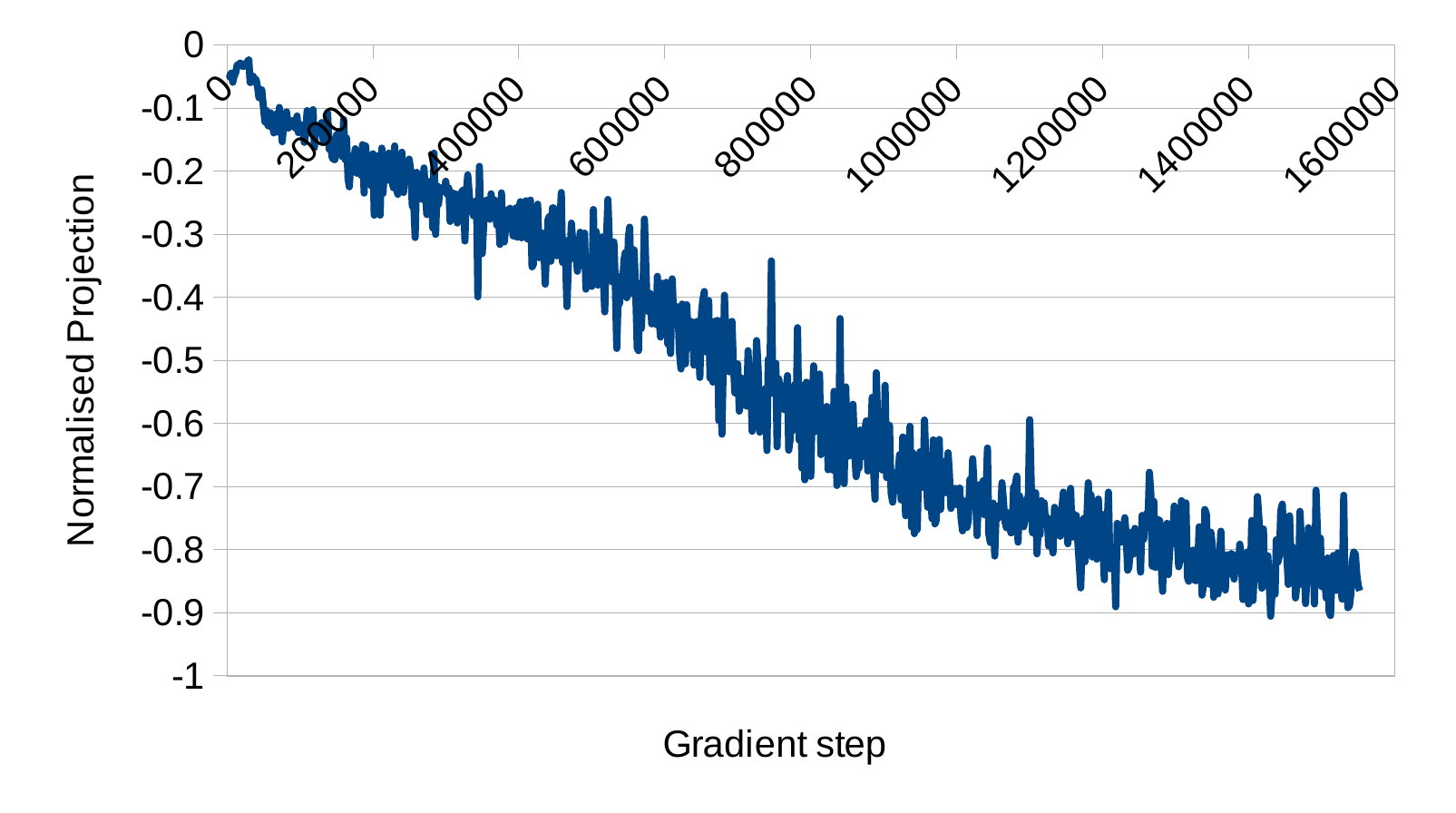}
		\caption{Beginning from a
		randomly-initialised GAN.}
		\label{local-conflict-from-start}
	\end{subfigure}
	\hfill
	\begin{subfigure}[b]{0.49\linewidth}
		\centering
		\includegraphics[width=1\linewidth]{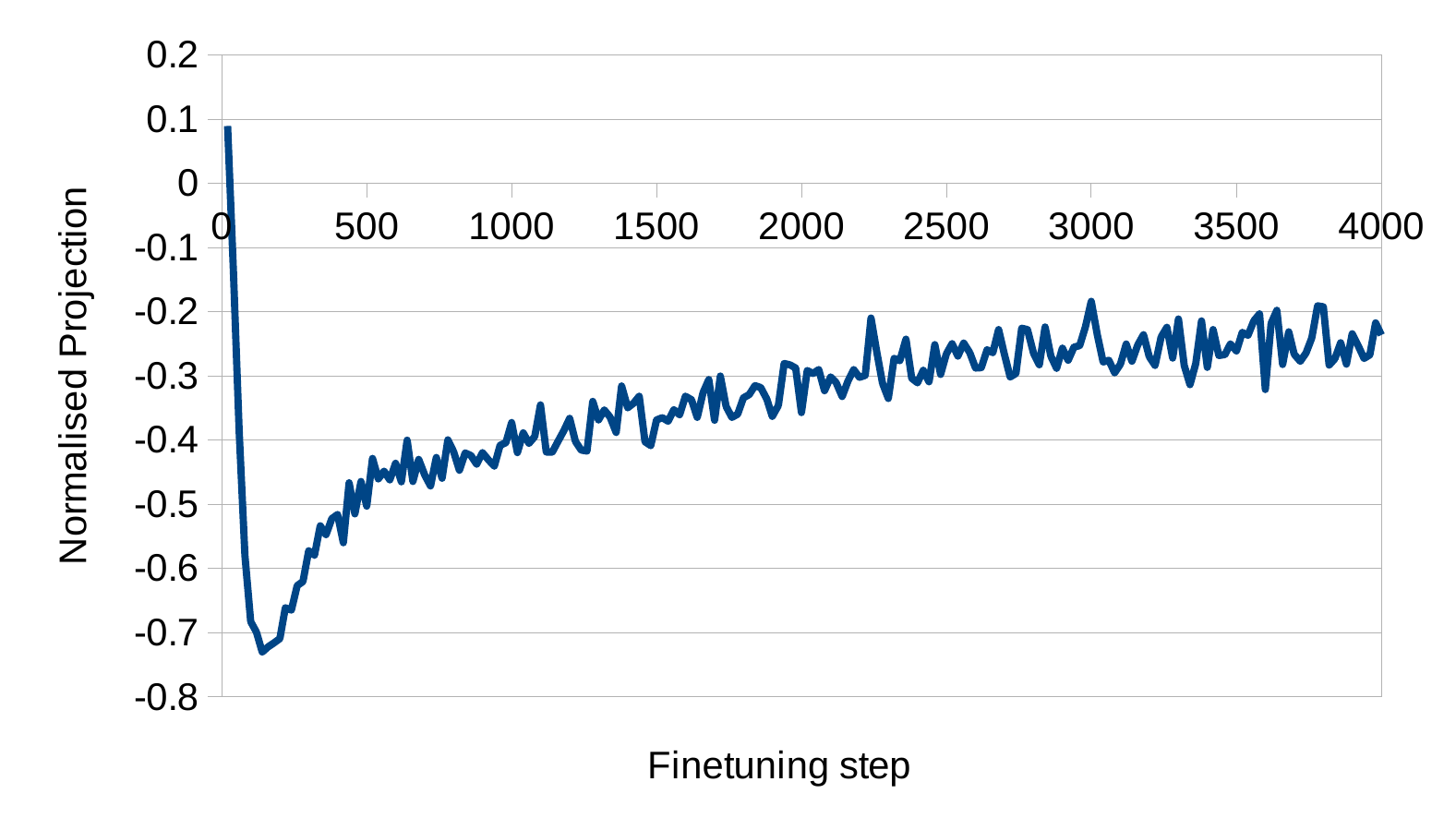}
		\caption{Adversarially finetuning
			a pretrained GAN.}
		\label{local-conflict-from-pretrained}
	\end{subfigure}
	\caption{Projecting normalised
		gradient vectors from
		$l_\textit{ordinary}$ and $l_\textit{(un)targeted}$
		onto one another.
	}
	\label{local-conflict-graph}
\end{figure}

\paragraph{Realistic pretraining}
It is widely accepted
that real image data occupy a
relatively low-dimensional and
contiguous manifold
 \citep[p.~160]{DBLP:books/daglib/0040158}.
Conversely, we know that adversarial
examples pervade the full input space:
it appears that
there is an adversarial example nearby
nearly any point in the input space.
Therefore,
a generator that is pretrained
using
only $l_\textit{ordinary}$
before
\textit{adversarially finetuning}
 by introducing our
additional loss term is
more successful than
using both loss terms from
a random initialisation.
By beginning our
search for unrestricted adversarial examples
in regions of realistic examples,
it is more likely that
there are global optima
of realistic adversarial examples
nearby.
Besides the generated images being
subjectively better,
Figure~\ref{local-conflict-from-pretrained}
shows that the gradients conflict to a
much lesser extent.
Note that \emph{any} existing GAN
architecture, pretrained checkpoint
and training algorithm could
be used here, allowing our
method to leverage the
significant advances being made
in this area.

\paragraph{Amalgamation of loss terms}
The most na\"{i}ve approach to jointly
optimising an ordinary GAN loss term
$l_\textit{ordinary}$
with our additional loss term $l_\textit{(un)targeted}$
is
to simply minimise their sum.
Part of the problem is that both
terms continue to be minimised
even if either one is `good enough';
the generator always aims to make
a misclassified example more strongly 
misclassified, which is not desirable.
We instead use the following per-example loss
term:
\[
l_\textit{finetune} = s(l_\textit{ordinary}) \cdot s(l_\textit{(un)targeted} - \kappa),\, \text{~where~}
s(l) =  
\begin{cases*}
1+\exp(l) & if $l \leq$ 0, \\
2+l  & otherwise.
\end{cases*}
\]
Here, $\kappa$ is a hyperparameter similar
to that in the
\citet{DBLP:conf/sp/Carlini017} attack:
it controls the confidence of the generated adversarial
examples.
If the difference between the desired logit and the
next-greatest logit is less than $\kappa$,
the generator is linearly rewarded for improving this
gap (gaining confidence); beyond a difference of $\kappa$ (once an 
example is `good enough'), the reward
exponentially decreases.
We use $\kappa = 0$ for our experiments.

\paragraph{Stochastic loss selection}
The gradients from the two loss terms are 
in conflict, and so
there is a danger that
one of these may dominate the other.
Unfortunately, this does
occur in practice:
the proportion of misclassified
generated inputs
rises quickly to almost 100\%, 
but the generated images were
noticeably unrealistic. This is a problem as
their correct label may change.
To address this, we introduce the `attack rate'~$\mu$.
During adversarial finetuning,
the finetuning loss term is used
at each step only
with probability $\mu$; with probability
$1-\mu$, the pretraining loss
($l_\textit{ordinary}$ only) is used.
These gradient steps facilitate the
optimisation of $l_\textit{ordinary}$
which otherwise is hindered by the
dominating gradient from
$l_\textit{(un)targeted}$.
As desired, this new hyperparameter allows the
success rate of the generated
unrestricted adversarial
examples to be traded off
with their realism.

\section{Experimental evaluation}
\label{experiments}

\begin{figure}[b!]

\setlength{\tabcolsep}{0.3pt} % whitespace between columns

\begin{minipage}[c]{0.48\linewidth}\centering

    \hspace{1cm} Target label 
    \vspace{2mm}

    \renewcommand{\arraystretch}{0.2} % whitespace between rows (multiplicative factor)
    \begin{tabular}{cr*{11}{c}}
    & & 0 & 1 & 2 & 3 & 4 & 5 & 6 & 7 & 8 & 9 & \hspace{0.7mm} None \vspace{1mm} \\
& 0 \hspace{2mm} & \hspace{15mm}& \includegraphics[width=4.8mm]{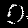} & \includegraphics[width=4.8mm]{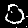} & \includegraphics[width=4.8mm]{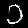} & \includegraphics[width=4.8mm]{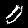} & \includegraphics[width=4.8mm]{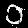} & \includegraphics[width=4.8mm]{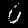} & \includegraphics[width=4.8mm]{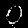} & \includegraphics[width=4.8mm]{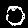} & \includegraphics[width=4.8mm]{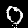} & \includegraphics[width=4.8mm]{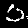} \\
& 1 \hspace{2mm} & \hspace{15mm}& \hspace{15mm}& \includegraphics[width=4.8mm]{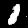} & \includegraphics[width=4.8mm]{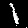} & \includegraphics[width=4.8mm]{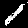} & \includegraphics[width=4.8mm]{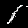} & \includegraphics[width=4.8mm]{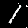} & \includegraphics[width=4.8mm]{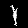} & \includegraphics[width=4.8mm]{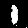} & \includegraphics[width=4.8mm]{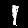} & \includegraphics[width=4.8mm]{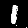} \\
\parbox[t]{4mm}{\multirow{11}{*}{\rotatebox[origin=c]{90}{Intended true label}}} & 2 \hspace{2mm} & \includegraphics[width=4.8mm]{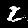} & \includegraphics[width=4.8mm]{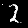} & \hspace{15mm}& \includegraphics[width=4.8mm]{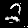} & \includegraphics[width=4.8mm]{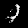} & \includegraphics[width=4.8mm]{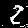} & \includegraphics[width=4.8mm]{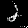} & \includegraphics[width=4.8mm]{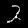} & \includegraphics[width=4.8mm]{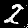} & \includegraphics[width=4.8mm]{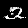} & \includegraphics[width=4.8mm]{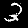} \\
& 3 \hspace{2mm} & \includegraphics[width=4.8mm]{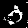} & \includegraphics[width=4.8mm]{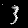} & \includegraphics[width=4.8mm]{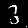} & \hspace{15mm}& \includegraphics[width=4.8mm]{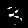} & \includegraphics[width=4.8mm]{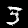} & \includegraphics[width=4.8mm]{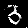} & \includegraphics[width=4.8mm]{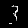} & \includegraphics[width=4.8mm]{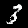} & \includegraphics[width=4.8mm]{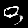} & \includegraphics[width=4.8mm]{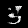} \\
& 4 \hspace{2mm} & \includegraphics[width=4.8mm]{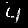} & \includegraphics[width=4.8mm]{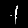} & \includegraphics[width=4.8mm]{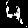} & \includegraphics[width=4.8mm]{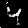} & \hspace{15mm}& \includegraphics[width=4.8mm]{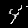} & \includegraphics[width=4.8mm]{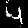} & \includegraphics[width=4.8mm]{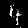} & \includegraphics[width=4.8mm]{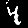} & \includegraphics[width=4.8mm]{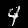} & \includegraphics[width=4.8mm]{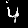} \\
& 5 \hspace{2mm} & \includegraphics[width=4.8mm]{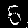} & \includegraphics[width=4.8mm]{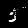} & \includegraphics[width=4.8mm]{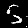} & \includegraphics[width=4.8mm]{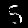} & \includegraphics[width=4.8mm]{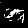} & \hspace{15mm}& \includegraphics[width=4.8mm]{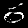} & \includegraphics[width=4.8mm]{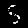} & \includegraphics[width=4.8mm]{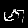} & \includegraphics[width=4.8mm]{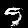} & \includegraphics[width=4.8mm]{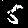} \\
& 6 \hspace{2mm} & \includegraphics[width=4.8mm]{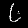} & \includegraphics[width=4.8mm]{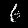} & \includegraphics[width=4.8mm]{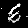} & \includegraphics[width=4.8mm]{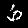} & \includegraphics[width=4.8mm]{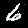} & \includegraphics[width=4.8mm]{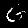} & \hspace{15mm}& \includegraphics[width=4.8mm]{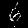} & \includegraphics[width=4.8mm]{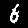} & \includegraphics[width=4.8mm]{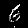} & \includegraphics[width=4.8mm]{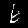} \\
& 7 \hspace{2mm} & \includegraphics[width=4.8mm]{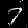} & \includegraphics[width=4.8mm]{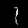} & \includegraphics[width=4.8mm]{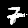} & \includegraphics[width=4.8mm]{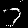} & \includegraphics[width=4.8mm]{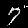} & \includegraphics[width=4.8mm]{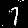} & \includegraphics[width=4.8mm]{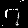} & \hspace{15mm}& \includegraphics[width=4.8mm]{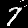} & \includegraphics[width=4.8mm]{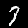} & \includegraphics[width=4.8mm]{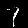} \\
& 8 \hspace{2mm} & \includegraphics[width=4.8mm]{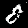} & \includegraphics[width=4.8mm]{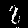} & \includegraphics[width=4.8mm]{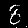} & \includegraphics[width=4.8mm]{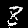} & \includegraphics[width=4.8mm]{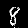} & \includegraphics[width=4.8mm]{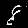} & \includegraphics[width=4.8mm]{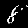} & \includegraphics[width=4.8mm]{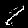} & \hspace{15mm}& \includegraphics[width=4.8mm]{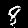} & \includegraphics[width=4.8mm]{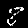} \\
& 9 \hspace{2mm} & \includegraphics[width=4.8mm]{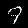} & \includegraphics[width=4.8mm]{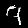} & \includegraphics[width=4.8mm]{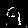} & \includegraphics[width=4.8mm]{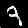} & \includegraphics[width=4.8mm]{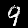} & \includegraphics[width=4.8mm]{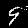} & \includegraphics[width=4.8mm]{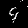} & \includegraphics[width=4.8mm]{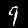} & \includegraphics[width=4.8mm]{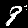} & \hspace{15mm}& \includegraphics[width=4.8mm]{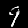} \\
    \end{tabular}
    \caption{Randomly-selected
    images generated by a GAN finetuned
    to attack \citeauthor{DBLP:conf/icml/WongK18}'s
    (\citeyear{DBLP:conf/icml/WongK18}) classifier,
    which is robust to perturbations.} 
    \label{generated-samples-wk}
\end{minipage}
\hfill{}
\begin{minipage}[c]{0.48\linewidth}
	\centering
	\includegraphics[width=1.02\linewidth]{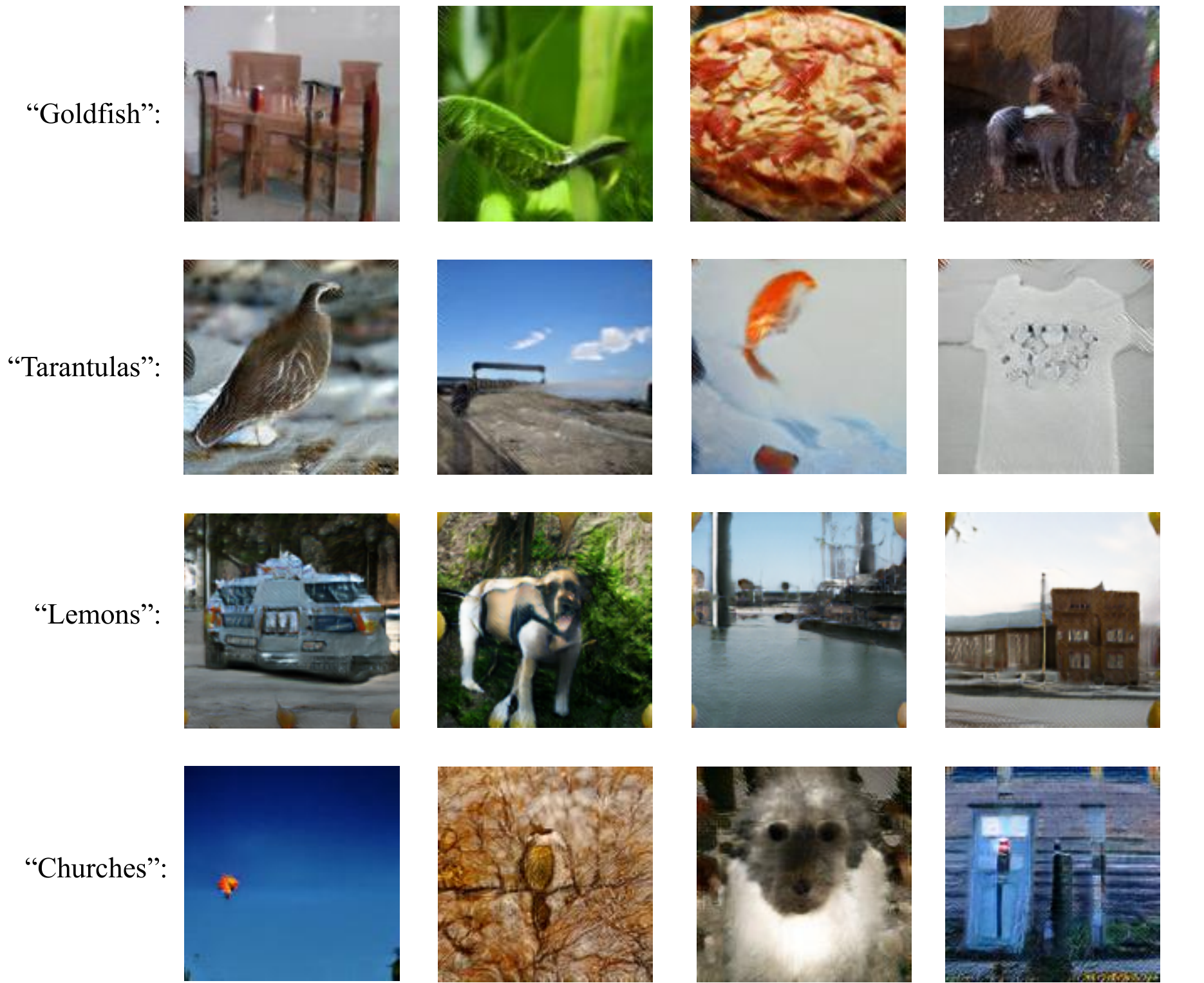}	
	\caption{Selected successful
	 targeted unrestricted adversarial examples
 	 on ImageNet, generated by a BigGAN
 	  \citep{DBLP:conf/iclr/BrockDS19}
  finetuned to attack ResNet-152
  \citep{DBLP:conf/cvpr/HeZRS16}.}
	\label{imagenet-figure}
\end{minipage}

\end{figure}

Our method aims to generate unrestricted
adversarial inputs in a way that adapts to the 
targeted classifier. We therefore conducted 
experiments to check whether the generated 
examples were in fact unrestricted, adversarial, 
realistic, and adapting to the classifier.
We then address some questions regarding the
performance and generality of our approach.

The MNIST dataset \citep{mnist-webpage}
is the main focus of the experimental evaluation,
because this is the most challenging domain
for the generation of realistic adversarial inputs.
State-of-the-art classifiers
perform very well, with around
0.2\% test error
\citep{DBLP:journals/corr/abs-1805-01890,
	DBLP:conf/icml/WanZZLF13}.
In particular,
attempts to create robust classifiers
have also been most successful on this dataset,
perhaps due to its simplicity
\citep{DBLP:journals/corr/abs-1809-02104}.
We target five pretrained classifiers 
provably robust to adversarial
perturbations:
there is guaranteed to be
no adversarial input within a distance $\epsilon$
of $p$\% of test inputs under the $l_\infty$ norm.
All five are the current state-of-the-art
in this domain, trained by
\cite{DBLP:conf/icml/WongK18},
and \cite{DBLP:journals/corr/abs-1811-02625}.
See Appendix~\ref{classifier-details} for details.

In~our experiments, we combine three
well-established generator architectures:
a Wasserstein GAN with gradient penalty (WGAN-GP)
\citep{DBLP:conf/nips/GulrajaniAADC17},
a conditional GAN 
\citep{DBLP:journals/corr/MirzaO14}
and an auxiliary classifier
GAN
\citep{DBLP:conf/icml/OdenaOS17}.
The generator is a convolutional
neural network, conditioned on class label.
The discriminator is a convolutional
neural network with two separate, diverging
final dense layers:
one acts as a conditional WGAN-GP critic, the
other as
an auxiliary classifier.
The auxiliary classifier helps the training converge,
but is not necessary.
Full details are given in Appendix~\ref{architectures-hyps}.

For each of the ten possible target labels---plus
the untargeted case, which aims
for any misclassification---a
GAN was adversarially finetuned.
After training converged,
the generators were used to produce
examples for all intended true labels,
which were then filtered so that the computed
label matched the target.
Images were generated until
200 such filtered examples
were generated for each
intended true label/target label
pairing or until 100 seconds had elapsed.
Interestingly, this led to no
adversarial examples with
intended true label `0'
and target classification `1',
so this case is omitted.
Figure~\ref{generated-samples-wk}
and Appendix~\ref{appendix-generated-samples}
give examples of generated
images for which the
computed label matches the
target classification.

\subsection{Efficacy of Attacks}
We claim that our method generates
unrestricted adversarial examples, which
are somewhat realistic. We empirically
verify each claim in turn.

Since our method does not work by
perturbing existing data, only a simple
sanity check was required to verify
that the generated images are not
close to images in the training set,
as could be caused by over-fitting.
We selected ten generated inputs that are
visually similar to the training set,
and computed the shortest distances between 
the images and all images in the
training set. The selected images
are given in Figure~\ref{realistic-examples}.
Table~\ref{perturbation-sizes} shows
that they are much further from any
training example 
than would be the case with a perturbation-based attack.

\begin{table}[]
\begin{minipage}[t]{0.55\linewidth}
	\setlength{\tabcolsep}{6pt} % whitespace between columns
	\begin{center}
		\caption{Comparison of typical
		perturbation magnitudes from the
		literature and ours.}
		\label{perturbation-sizes}
		\begin{tabular}{>{\centering\arraybackslash}p{0.7cm}>{\centering\arraybackslash}p{1.45cm}>{\centering\arraybackslash}p{4cm}}
			\toprule
			{Metric} & {Nearest neighbour seen} &
			Typical perturbation magnitude\\
			\midrule
			$l_0$ & 508\hphantom{.000} &
			 <40
			 \citep{DBLP:journals/corr/abs-1804-05805}\\
			 $l_1$ & \mbox{\hphantom{0}22.8\hphantom{00}} & <5
			 \citep{DBLP:conf/dsn/LuCCY18} \\
			 $l_2$ & \mbox{\hphantom{00}3.28\hphantom{0}} & \textasciitilde1.5
			 \citep{DBLP:journals/corr/abs-1805-09190} \\
			 $l_\infty$ & \mbox{\hphantom{00}0.838} &  \textasciitilde0.1
			 \citep{DBLP:conf/icml/WongK18} \\
			\bottomrule
		\end{tabular}
	\end{center}
\end{minipage}
\hfill{}
\begin{minipage}[t]{0.4\linewidth}
	\caption{Ten selected unrestricted
		adversarial inputs used for 
		Table \ref{perturbation-sizes}.}
	\centering%
	\includegraphics{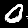}
	\includegraphics{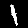}
	\includegraphics{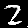}
	\includegraphics{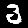}
	\includegraphics{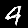}
	\\ \vspace{1mm}
	\includegraphics{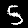}
	\includegraphics{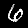}
	\includegraphics{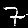}
	\includegraphics{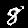}
	\includegraphics{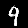}
	\label{realistic-examples}
\end{minipage}
\end{table}

\label{are-examples-adversarial}
Next, we evaluate the whether our
method is successful in generating
adversarial inputs.
Our method generates such inputs
if and only if
their true label matches their
intended true label
given as input to the generator, else
the generator could simply be ignoring
this input and generating images which
visually match the target class.
To check that the finetuned
generators are behaving
as hoped,
we used workers on Amazon's MTurk
platform to classify the generated images.
For cost reasons,
we only carried out the MTurk experiments
targeting Wong and Kolter's
provably-robust network
\citep{DBLP:conf/icml/WongK18}, not
any of the MixTrain models.
We used a sample size of 100 judges
for each intended true label/target label
pair for each experiment.
Figure~\ref{labelme-mnist-finetuned-robust}
shows the proportion of inputs for which
not only does the
label computed by the
classifier match the target label,
but the human-judged true label
matches the intended true label
specified to the generator.
The mean number of correct labels
for the untargeted attack is
80\%.
This can be considered to be
the success rate of our attack.

\begin{figure}
    \begin{minipage}[t]{0.480\linewidth}\centering
    \hspace{1cm} Target label 

    \begin{tabular}{rr*{11}{L}}
    
 & & $0$ & $1$ & $2$ & $3$ & $4$ & $5$ & $6$ & $7$ & $8$ & $9$ & None \\
\parbox[t]{-9cm}{\multirow{11}{*}{\rotatebox[origin=c]{90}{Intended true label}}} & 0 \hspace{0.8mm} & & 96  & 94  & 90  & 85  & 96  & 97  & 99  & 85  & 89  & 95  \\
& 1 \hspace{0.8mm} & XX  & & 66  & 88  & 69  & 97  & 89  & 74  & 91  & 81  & 87  \\
& 2 \hspace{0.8mm} & 69  & 89  & & 82  & 58  & 82  & 70  & 64  & 79  & 49  & 75  \\
& 3 \hspace{0.8mm} & 43  & 84  & 81  & & 68  & 74  & 46  & 82  & 54  & 71  & 53  \\
& 4 \hspace{0.8mm} & 84  & 67  & 86  & 74  & & 75  & 96  & 79  & 82  & 77  & 76  \\
& 5 \hspace{0.8mm} & 58  & 75  & 70  & 78  & 79  & & 52  & 82  & 69  & 81  & 75  \\
& 6 \hspace{0.8mm} & 82  & 90  & 95  & 73  & 84  & 84  & & 86  & 94  & 84  & 82  \\
& 7 \hspace{0.8mm} & 75  & 75  & 88  & 82  & 76  & 95  & 88  & & 92  & 59  & 80  \\
& 8 \hspace{0.8mm} & 76  & 85  & 91  & 76  & 98  & 97  & 77  & 75  & & 91  & 83  \\
& 9 \hspace{0.8mm} & 77  & 68  & 90  & 84  & 95  & 92  & 88  & 95  & 95  & & 90  \\
 &  &  &  &  &  &  &  &  &  &  & \\[-3mm]
& Mean \hspace{0.6mm} & 70 & 81 & 85 & 81 & 79 & 88 & 78 & 82 & 82 & 76 & 80\\\end{tabular}
 \caption{The success rates of the adversarial attacks by finetuned GANs (the computed label matches the target label and the true label remains the same).} 
\label{labelme-mnist-finetuned-robust}
\end{minipage}
\hfill{}
    \begin{minipage}[t]{0.49\linewidth}\centering
    \hspace{1cm} Target label 

    \begin{tabular}{rr*{11}{O}}
    
 & & $0$ & $1$ & $2$ & $3$ & $4$ & $5$ & $6$ & $7$ & $8$ & $9$ & None \\
\parbox[t]{-9cm}{\multirow{11}{*}{\rotatebox[origin=c]{90}{Intended true label}}} & 0 \hspace{0.8mm} & & 40  & 60  & 56  & 34  & 46  & 51  & 40  & 36  & 63  & 51  \\
& 1 \hspace{0.8mm} & XX  & & 37  & 52  & 36  & 51  & 81  & 40  & 53  & 35  & 49  \\
& 2 \hspace{0.8mm} & 30  & 37  & & 43  & 40  & 42  & 35  & 37  & 55  & 32  & 54  \\
& 3 \hspace{0.8mm} & 39  & 39  & 43  & & 34  & 40  & 40  & 42  & 45  & 48  & 40  \\
& 4 \hspace{0.8mm} & 51  & 50  & 34  & 38  & & 37  & 46  & 42  & 41  & 43  & 40  \\
& 5 \hspace{0.8mm} & 32  & 34  & 32  & 36  & 43  & & 42  & 36  & 37  & 55  & 51  \\
& 6 \hspace{0.8mm} & 51  & 39  & 45  & 36  & 57  & 46  & & 45  & 57  & 40  & 46  \\
& 7 \hspace{0.8mm} & 47  & 48  & 53  & 33  & 42  & 58  & 41  & & 52  & 44  & 39  \\
& 8 \hspace{0.8mm} & 29  & 46  & 47  & 55  & 44  & 48  & 36  & 39  & & 42  & 60  \\
& 9 \hspace{0.8mm} & 38  & 34  & 50  & 49  & 54  & 53  & 53  & 69  & 57  & & 67  \\
 &  &  &  &  &  &  &  &  &  &  & \\[-3mm]
& Mean \hspace{0.6mm} & 40 & 41 & 45 & 44 & 43 & 47 & 47 & 43 & 48 & 45 & 50\\\end{tabular}
 \caption{How often adversarial images are not identified as being generated. If the generated images were completely realistic, the expected result would be 90.} 
\label{oddoneout-mnist-finetuned-robust}
\end{minipage}
\end{figure}

% Generated by Python file from JSON of results.
% To generate, run "python create_results_tables.py"

\label{realistic-experiments}

We now investigate if the
generated examples are realistic.
A set of inputs is \textit{realistic} with respect to a
dataset if a human cannot reliably identify to which set
an example belongs.
To check this,
we again used MTurk workers.
After familiarising themselves with examples
from the training dataset,
each worker had to pick which image
out of ten was
most likely to have been
generated.
Figure~\ref{oddoneout-mnist-finetuned-robust}
shows the proportion of the time
that generated images were not identified
as such.

\label{how-does-performance-depend-on-the-target-network}
For comparison, we
repeated these experiments 
but attacking a non-robust classifier network.
The untargeted success rate
was 90\% (vs.~80\% 
against the robust classifier),
and 60\% (vs.~50\%) were not identified
as being generated.
Similar differences were seen in targeted
attacks;
see Appendix~\ref{nonrobust-results}. 

\subsection{Adaptivity to Adversarial Training Defences}
\label{adv-train-exp}

In the above experiments we
have evaluated our method against 
pretrained classifiers that are
provably robust to 
adversarial perturbations.
We now investigate whether standard adversarial
training \citep{DBLP:conf/iclr/MadryMSTV18}
against our attack in particular
is effective.
Starting with a pretrained GAN
and classifier,
we iterate `training rounds' consisting of two
phases.
First, a GAN is adversarially finetuned
(starting from the pretrained GAN each time)
for a fixed period
to attack the classifier. Second,
80,000 generated unrestricted adversarial examples
are added to the existing training dataset,
and the classifier
\emph{continues} training until almost 100\%
accuracy is achieved.

\begin{figure}[]
	\centering
	\begin{subfigure}[t]{0.32\linewidth}
		\centering
		\includegraphics[width=\linewidth]{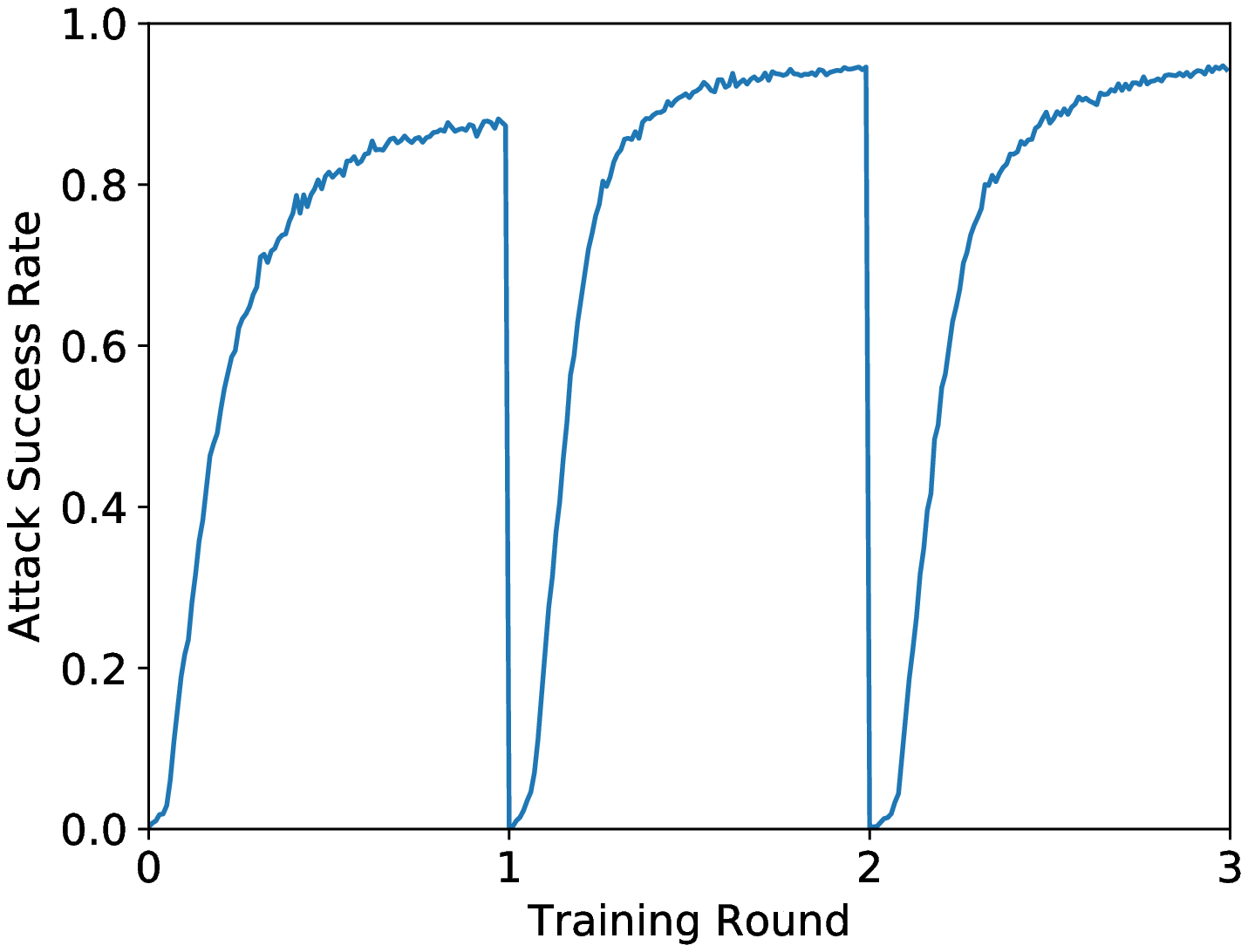}
		\caption{Adversarial training against our attack, first few training rounds.}
		\label{adv-train-beg}
	\end{subfigure}
	\hfill
	\begin{subfigure}[t]{0.32\linewidth}
		\centering
		\includegraphics[width=\linewidth]{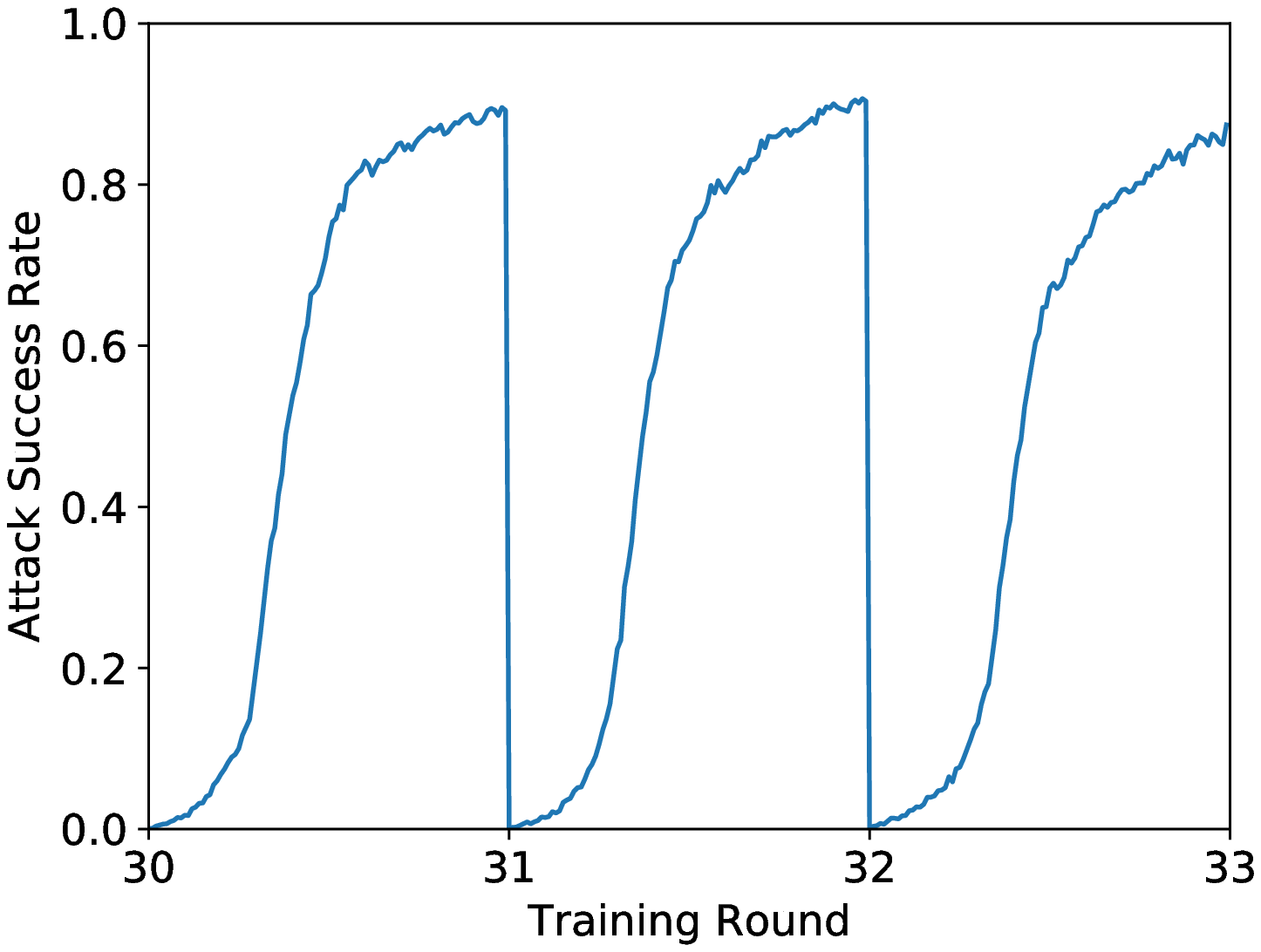}
		\caption{Adversarial training against our attack, later training rounds.}
		\label{adv-train-end}
	\end{subfigure}
	\hfill
	\begin{subfigure}[t]{0.32\linewidth}
		\centering
		\includegraphics[width=\linewidth]{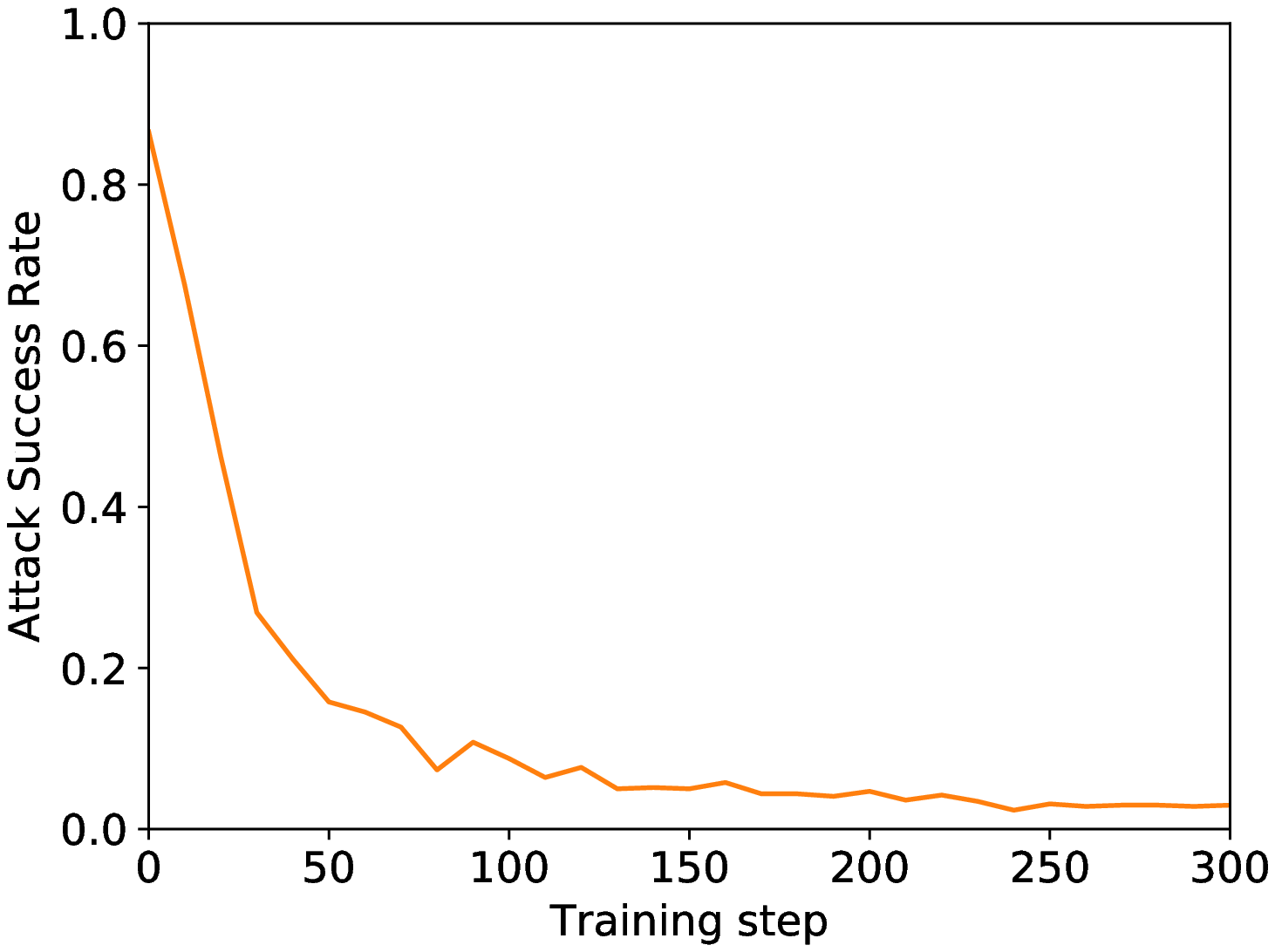}
		\caption{Adversarial training against \cite{DBLP:conf/nips/SongSKE18}.}
		\label{adv-train-songetal}
	\end{subfigure}
	\caption{Plots showing
		attack efficacy
		in the presence of adversarial training.
	}
\end{figure}

Figure~\ref{adv-train-beg} shows that,
for the first few training rounds, adversarial
finetuning is successful: the proportion of
examples generated which fool the
classifier increases to over 80\%.
Figure~\ref{adv-train-end}
shows the same story
30 rounds (and hence hundreds of thousands
of classifier gradient steps) in.
Although the classifier may be able
to defend against the kinds of attacks
learnt by the generator in previous training
rounds, the generator's opportunity to
adversarially finetune again allows it
to generate adversarial examples of a kind
not seen before by the classifier.
Since the generator is unrestricted,
it seems unlikely to `run out' of these.
For more details on these experiments, 
see Appendix~\ref{adv-train-exp-appendix}.

\subsection{Scaling to ImageNet}
\label{scale-exp}

While the MNIST classifiers are the most
challenging to fool, MNIST is a relatively
small and simple dataset.
To demonstrate the scalability of our method,
we apply it to the notoriously large and
complex ImageNet-1K dataset.
We also take advantage of the fact
that our method works
with \emph{any} pretrained GAN by using
the author's `officially unofficial'
published code and checkpoints for the
current state-of-the-art, BigGAN \citep{DBLP:conf/iclr/BrockDS19}.
In the untargeted case, our method is able to finetune this
BigGAN to fool the classifier >99\%
of the time within 40 gradient steps
(compared to the $10^5$ taken to
train from scratch).
Our main focus, though, is on the much more challenging
targeted attack.
We adversarially
finetuned a BigGAN several times, selecting a variety
of target classes.
We found that typically, on the order of 100 gradient
steps were required for >10\% of generated examples
to be classified (top-1) as the target class.
Compared to MNIST,
each ImageNet gradient step takes
about 100x longer to compute,
but the 100x decrease in the number
of gradient steps required compensates for this,
resulting in a similar compute time overall.
Image quality as measured by Inception Score \citep{DBLP:conf/nips/SalimansGZCRCC16}
typically
decreased from 70, which is
slightly better than mid-2018
 state-of-the-art of 52
\citep{DBLP:conf/icml/ZhangGMO19}
to the mid-2017 state-of-the-art (WGAN-GP) of 12 
\citep{DBLP:conf/eccv/ShmelkovSA18}.
We speculate that if the GAN were finetuned for
significantly longer, the gradient from
the discriminator would learn to regain some
of this lost realism.
Figure~\ref{imagenet-figure} shows selected
samples of generated adversarial examples;
Appendix \ref{imagenet-appendix} has a more extensive collection.

\subsection{Ablation Study}
To determine the contribution
of our method, the MNIST experiments described were
rerun but using a GAN not adversarially finetuned
as a baseline.
Unsurprisingly,
the desired misclassifications occurred
vastly less frequently
than when generated by a finetuned GAN.
Furthermore, of those
which were misclassified,
the proportion
for which
the true label also matched the
intended true label
was also significantly
lower without adversarial finetuning:
66\% for untargeted attacks and
58\% on average for targeted attacks,
compared to 80\% for both categories
after finetuning.
Full results are given
in Appendix~\ref{pretrained-results}.
In short,
we found that
finetuning a GAN using our method
roughly maintains how realistic its
generated expected-adversarial images
are, while significantly increasing their quantity
and increasing
the attack success rate by around 15 percentage points.

\subsection{Threats to Validity}

The evaluation of the success
of the attacks relies
relies on data
provided by the MTurk workers.
We therefore employed measures
to safeguard the quality of this
data,
described in Appendix~\ref{mturk-app}.
We also believe that our method will generalise
to any dataset and domain for which GANs
can be trained successfully. However,
this has only been demonstrated
on two image classification tasks
(albeit dissimilar in nature).
Lastly, intuition suggests that our method
will be able to adapt to find
unrestricted adversarial examples
for almost any defence method,
since it is so free to generate
inputs without the constraints
that current defence methods
rely upon. However, we have only
demonstrated it explicitly for the most
popular standard defence;
future work may find a defence
against our approach.

\section{Related Work}
\subsection{Comparison to the State of the Art}

We compare our method to that of
\cite{DBLP:conf/nips/SongSKE18}, the current state of the art 
in  generating unrestricted adversarial examples.
Like ours, this method leverages a pretrained GAN.
It differs, however, in how adversarial examples are then 
produced.
Instead of adversarially finetuning
the generator, it searches
for an input to the generator
that both deceives the target network 
and are confidently correctly classified by the
discriminator's auxiliary classifier
(an ACGAN \citep{DBLP:conf/icml/OdenaOS17} is
required in this case).
The GAN training 
is therefore blind to the 
target classifier.

Our model achieves similar success rates in
generating unrestricted 
adversarial examples:
our success rate of 80\%
(cf. Section~\ref{are-examples-adversarial})
is roughly comparable to
that of
\citet{DBLP:conf/nips/SongSKE18}, 88.8\%.
For comparison, we repeated the realism
experiments from Section \ref{realistic-experiments}, 
with the difference that judges
were asked to identify the one generated
image from a choice of two.
In this case,
\citeauthor{DBLP:conf/nips/SongSKE18} report
that participants select
the generated image as the more realistic
21.8\% of the time while
for our untargeted attack, this figure is 24\%;
completely realistic image would be
chosen 50\% of the time.
Full results are given in
Appendix~\ref{abtest-results}.

Beyond achieving comparable attack
success rates, our approach has four significant  
advantages over prior work.
Firstly: adaptivity.
In Section \ref{adv-train-exp} we have shown that 
our model is capable of iteratively adapting to an adversarially
trained classifier. 
By contrast, \citeauthor{DBLP:conf/nips/SongSKE18}'s method 
performs poorly against adversarial 
training because the GAN is
not trained with respect to a target classifier,
remaining fixed after the attack begins.
Therefore,
if a classifier learns
to be correct in the space their algorithm searches, it 
will no longer be able to generate images
different enough
to be adversarial.
Figure \ref{adv-train-songetal}
shows that standard adversarial training
quickly and effectively defends against
\citeauthor{DBLP:conf/nips/SongSKE18}'s attack,
while it fails against ours.
Secondly: efficiency.
Once trained, our method 
requires only a single forward pass to
generate adversarial examples. 
\citeauthor{DBLP:conf/nips/SongSKE18}~require 100--500 iterations, each with
forward and backward passes through both the 
generator and classifier. Our method
is therefore 400--2,000$\times$ more efficient.
Lastly: scale and versatility. 
Section \ref{scale-exp} shows
that our model scales to ImageNet, a 
dataset with
dimensionality 16$\times$ greater
than the largest \citeauthor{DBLP:conf/nips/SongSKE18} demonstrate on. Our method
has the further benefit that we can use
any pretrained GAN, such as BigGAN \citep{DBLP:conf/iclr/BrockDS19}. 
\citeauthor{DBLP:conf/nips/SongSKE18} depend on 
an auxiliary classifier for larger datasets, 
which BigGAN does not provide.

\subsection{Other Related Work}

\cite{DBLP:journals/corr/abs-1904-07793}
independently propose
a method which is superficially similar to ours:
they also train a GAN to directly generate adversarial
examples. However, instead of using the
ordinary GAN loss to ensure that the
adversarial examples are sufficiently realistic,
they instead use a new loss term. This term,
$\|g_\textit{pretrained}(z) - g(z)\|_p$,
penalises the generator $g$
given input $z$ proportional to the
deviation of its output
from what it would have output
immediately before adversarial finetuning.
Our approach, to use the
ordinary GAN loss for this purpose,
allows for truly unrestricted adversarial
examples, giving the training procedure
much more scope to adapt to circumvent
any specific defences (such as
robustness to $l_p$ perturbations).
\citeauthor{DBLP:journals/corr/abs-1904-07793}'s
choice of loss term has the unfortunate effect
of preventing the generator from generating
either unrestricted adversarial examples
or examples which are sure to fall
within an $l_p$-norm ball of a realistic input.
Our method has three further advantages
over this work:
we evaluate against state-of-the-art
provably-robust networks rather than
ad-hoc classifiers;
we conduct a user study to
quantitatively verify the proportion
of generated adversarial examples
which maintain the correct label
rather than assuming that
this is 100\%, which is unlikely;
and we demonstrate that our approach
scales beyond MNIST (to ImageNet).

\cite{DBLP:journals/tissec/SharifBBR19}
train a network to generate
patterned spectacles, which, when added
to an image of a face, cause misclassification.
They also adapt this approach to generate
unrestricted adversarial examples for MNIST
using an approach quite similar to ours.
However, this only achieves a success
rate of 8.34\% against a classifier
which was state-of-the-art in 2017,
which is reduced to 0.83\% after filtering
to ``only the digits that
where likely to be comprehensible by humans''.
In contrast, we achieve around 80\% accuracy
against current state-of-the-art robust classifiers.

A wide range of work trains networks
to generate adversarial \emph{perturbations}
\citep{
	DBLP:conf/sp/HayesD18,
	DBLP:conf/aaai/BalujaF18,
	DBLP:conf/ijcai/XiaoLZHLS18,
	DBLP:journals/corr/abs-1811-12026,
	DBLP:conf/cvpr/PoursaeedKGB18}.
While these must also balance realism and 
adversarial success, the key difference is that 
we
generate \emph{unrestricted} adversarial examples,
allowing attacks to succeed when
constrained
perturbations provably fail.

\cite{DBLP:journals/corr/abs-1906-07920}
introduce a search for pairs of nearby
unrestricted
adversarial examples,
but unfortunately cannot ensure that
their true label is meaningful;
if the search starting point
is random, it is
overwhelmingly likely not to be.
If instead it is a known input,
the examples are not unrestricted.

\section{Conclusion}

We have introduced an algorithm which
trains a GAN to generate unrestricted
adversarial
inputs; we demonstrate that
these, as expected, are successful
against state-of-the-art classifiers robust to
perturbation attacks.
The key novelty in our attack procedure
is that it entails the
tuning of the weights of
the generator to target a specific
network.
As a result, it can be considered
\emph{adaptive}: we have shown that,
while prior work is quickly mitigated by 
standard adversarial training, 
our attack adapts to find a new 
way of fooling the classifier.
In addition, once the generator
is adversarially finetuned, it
becomes an endless supply 
of cheap adversarial examples:
generation of adversarial examples
requires a single forward pass rather
than execution of any optimisation algorithm,
resulting in a 400--2000$\times$ speedup over
the state of the art.
We have also demonstrated that any existing
GAN codebase can easily be used by adapting
BigGAN to generate unrestricted adversarial
examples for ImageNet.

\newpage

\bibliography{gans4tests}
\bibliographystyle{iclr2020_conference}

\newpage
\appendix

\section{Samples of ImageNet Unrestricted Adversarial Examples}
\label{imagenet-appendix}

Randomly-selected successful targeted
unrestricted adversarial examples
generated using
adversarially finetuned BigGANs
\citep{DBLP:conf/iclr/BrockDS19}.
The targeted classifier is
ResNet-152
\citep{DBLP:conf/cvpr/HeZRS16},
the highest-accuracy pretrained
classifier packaged with PyTorch.
Besides setting our attack rate at
0.1, all configuration and
hyperparameters are as described
in the BigGAN `officially unofficial'
codebase.\footnote{\url{https://github.com/ajbrock/BigGAN-PyTorch}}

\vspace{0.8cm}

\begin{figure}[h]
\begin{minipage}[t]{0.47\linewidth}
	\begin{tabularx}{\linewidth}{YYYY}
		\includegraphics[width=0.9\linewidth]{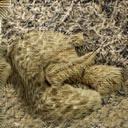} & \includegraphics[width=0.9\linewidth]{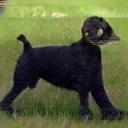}
		& \includegraphics[width=0.9\linewidth]{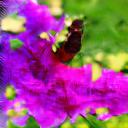} & \includegraphics[width=0.9\linewidth]{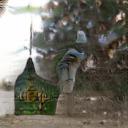} \\
		horned rattlesnake & curly-coated retriever & admiral & birdhouse\\
		\includegraphics[width=0.9\linewidth]{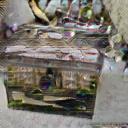} & \includegraphics[width=0.9\linewidth]{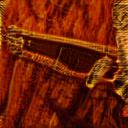}
		& \includegraphics[width=0.9\linewidth]{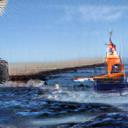} & \includegraphics[width=0.9\linewidth]{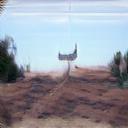} \\
		 chest & horn & lifeboat & solar dish \\
		 \includegraphics[width=0.9\linewidth]{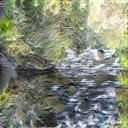} & \includegraphics[width=0.9\linewidth]{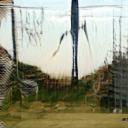}
		& \includegraphics[width=0.9\linewidth]{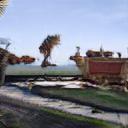} & \includegraphics[width=0.9\linewidth]{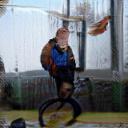} \\
		 stone wall & suspension bridge & thresher & unicycle\\
		 \includegraphics[width=0.9\linewidth]{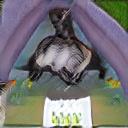} & \includegraphics[width=0.9\linewidth]{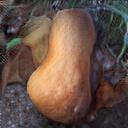}
		& \includegraphics[width=0.9\linewidth]{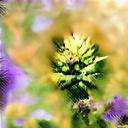} & \includegraphics[width=0.9\linewidth]{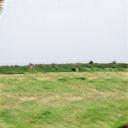} \\
		 comic book & butternut squash & cardoon & hay
	\end{tabularx}
	\caption{Successful targeted unrestricted
		adversarial examples
		for target class `tabby cat'.}
\end{minipage}\hfill%
\begin{minipage}[t]{0.47\linewidth}
	\begin{tabularx}{\linewidth}{YYYY}
		\includegraphics[width=0.9\linewidth]{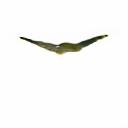} & \includegraphics[width=0.9\linewidth]{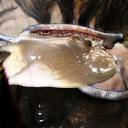}
		& \includegraphics[width=0.9\linewidth]{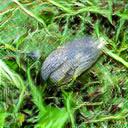} & \includegraphics[width=0.9\linewidth]{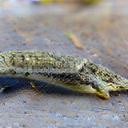} \\
		kite & spotted salamander & terrapin & alligator lizard \\
	   \includegraphics[width=0.9\linewidth]{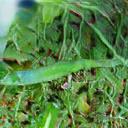} & \includegraphics[width=0.9\linewidth]{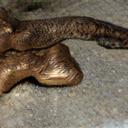}
		& \includegraphics[width=0.9\linewidth]{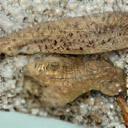} & \includegraphics[width=0.9\linewidth]{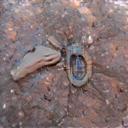} \\
		green lizard & night snake & horned rattlesnake & centipede\\
	   \includegraphics[width=0.9\linewidth]{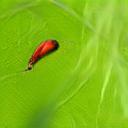} & \includegraphics[width=0.9\linewidth]{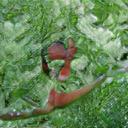}
		& \includegraphics[width=0.9\linewidth]{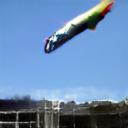} & \includegraphics[width=0.9\linewidth]{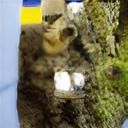} \\
		lady bug & howler monkey & airship & combination lock \\
	   \includegraphics[width=0.9\linewidth]{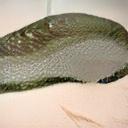} & \includegraphics[width=0.9\linewidth]{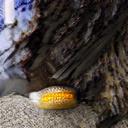}
		& \includegraphics[width=0.9\linewidth]{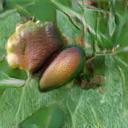} & \includegraphics[width=0.9\linewidth]{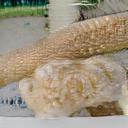} \\
		sombrero &  corn & acorn & capitulum
	\end{tabularx}
	\caption{Successful targeted unrestricted
		adversarial examples
		for target class `slug'.}
\end{minipage}

\vspace{0.5cm}

\begin{minipage}[t]{0.47\linewidth}
	\begin{tabularx}{\linewidth}{YYYY}
		\includegraphics[width=0.9\linewidth]{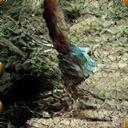} & \includegraphics[width=0.9\linewidth]{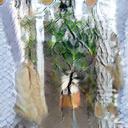}
		& \includegraphics[width=0.9\linewidth]{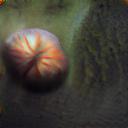} & \includegraphics[width=0.9\linewidth]{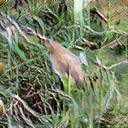} \\
		cock & black widow & nautilus & bittern \\
		\includegraphics[width=0.9\linewidth]{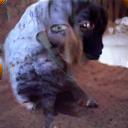} & \includegraphics[width=0.9\linewidth]{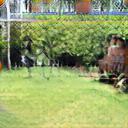}
		& \includegraphics[width=0.9\linewidth]{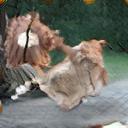} & \includegraphics[width=0.9\linewidth]{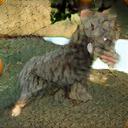} \\
		bluetick & english setter & sussex spaniel & briard \\
		\includegraphics[width=0.9\linewidth]{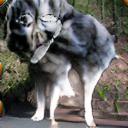} & \includegraphics[width=0.9\linewidth]{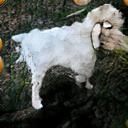}
		& \includegraphics[width=0.9\linewidth]{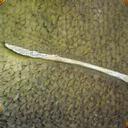} & \includegraphics[width=0.9\linewidth]{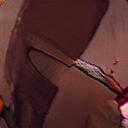} \\
		eskimo dog & standard poodle & ladle & mailbag \\
		\includegraphics[width=0.9\linewidth]{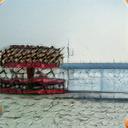} & \includegraphics[width=0.9\linewidth]{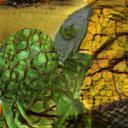}
		& \includegraphics[width=0.9\linewidth]{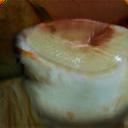} & \includegraphics[width=0.9\linewidth]{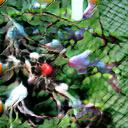} \\
		paddle wheel & custard apple & eggnog & conker 
	\end{tabularx}	
	\caption{Successful targeted unrestricted
		adversarial examples
		for target class `orange'.}
\end{minipage}\hfill%
\begin{minipage}[t]{0.47\linewidth}
	\begin{tabularx}{\linewidth}{YYYY}
		\includegraphics[width=0.9\linewidth]{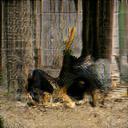} & \includegraphics[width=0.9\linewidth]{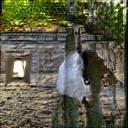}
		& \includegraphics[width=0.9\linewidth]{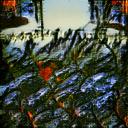} & \includegraphics[width=0.9\linewidth]{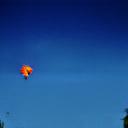} \\
		cock & dhole & squirrel monkey & balloon \\
		\includegraphics[width=0.9\linewidth]{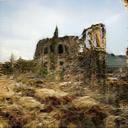} & \includegraphics[width=0.9\linewidth]{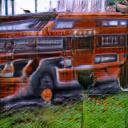}
		& \includegraphics[width=0.9\linewidth]{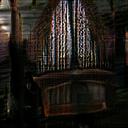} & \includegraphics[width=0.9\linewidth]{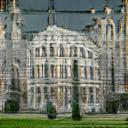} \\
		castle & garbage truck & organ & palace \\
		\includegraphics[width=0.9\linewidth]{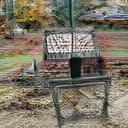} & \includegraphics[width=0.9\linewidth]{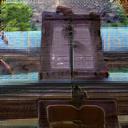}
		& \includegraphics[width=0.9\linewidth]{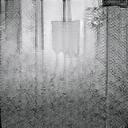} & \includegraphics[width=0.9\linewidth]{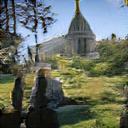} \\
		park bench & revolver & shower curtain & stupa \\
		\includegraphics[width=0.9\linewidth]{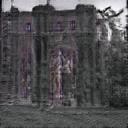} & \includegraphics[width=0.9\linewidth]{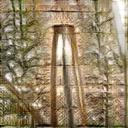}
		& \includegraphics[width=0.9\linewidth]{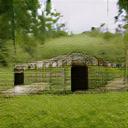} & \includegraphics[width=0.9\linewidth]{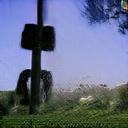} \\
		triumphal arch & water tower & yurt & traffic light \\
	\end{tabularx}
	\caption{Successful targeted unrestricted
		adversarial examples
		for target class `church'.}
\end{minipage}
\end{figure}

\newpage

\section{Samples of MNIST Unrestricted Adversarial Examples}
\label{appendix-generated-samples}

\vspace{0.8cm}

 \begin{figure}[H]
            \begin{subfigure}{0.45\textwidth}
                \includegraphics[width=0.95\linewidth, clip, trim=0.0cm 5.30cm 0.0cm 0.0cm]{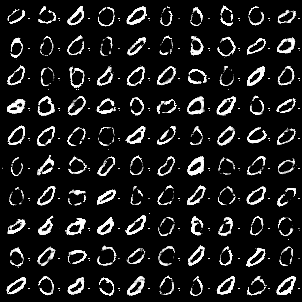}
                \caption{Intended true label `0'.} 
            \end{subfigure}
        \hspace*{\fill}
            \begin{subfigure}{0.45\textwidth}
                \includegraphics[width=0.95\linewidth, clip, trim=0.0cm 5.30cm 0.0cm 0.0cm]{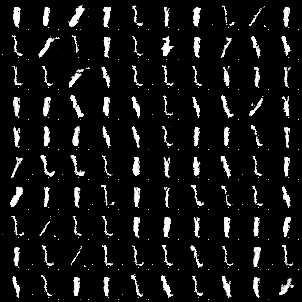}
                \caption{Intended true label `1'.} 
            \end{subfigure}
        \medskip
            \begin{subfigure}{0.45\textwidth}
                \includegraphics[width=0.95\linewidth, clip, trim=0.0cm 5.30cm 0.0cm 0.0cm]{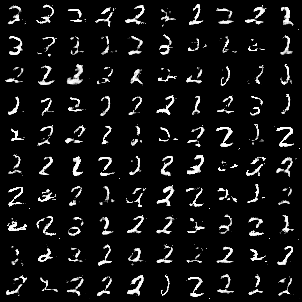}
                \caption{Intended true label `2'.} 
            \end{subfigure}
        \hspace*{\fill}
            \begin{subfigure}{0.45\textwidth}
                \includegraphics[width=0.95\linewidth, clip, trim=0.0cm 5.30cm 0.0cm 0.0cm]{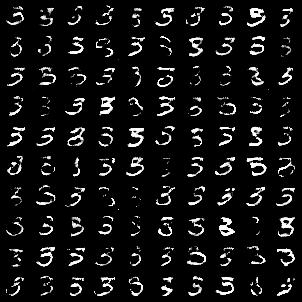}
                \caption{Intended true label `3'.} 
            \end{subfigure}
        \medskip
            \begin{subfigure}{0.45\textwidth}
                \includegraphics[width=0.95\linewidth, clip, trim=0.0cm 5.30cm 0.0cm 0.0cm]{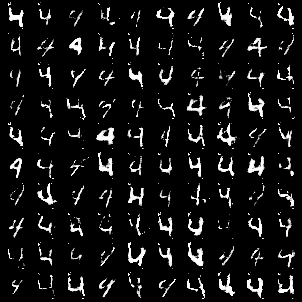}
                \caption{Intended true label `4'.} 
            \end{subfigure}
        \hspace*{\fill}
            \begin{subfigure}{0.45\textwidth}
                \includegraphics[width=0.95\linewidth, clip, trim=0.0cm 5.30cm 0.0cm 0.0cm]{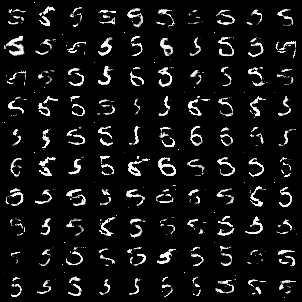}
                \caption{Intended true label `5'.} 
            \end{subfigure}
        \medskip
            \begin{subfigure}{0.45\textwidth}
                \includegraphics[width=0.95\linewidth, clip, trim=0.0cm 5.30cm 0.0cm 0.0cm]{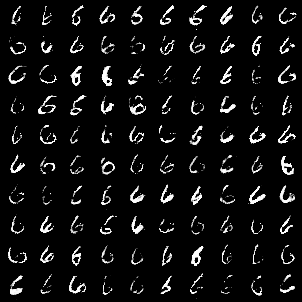}
                \caption{Intended true label `6'.} 
            \end{subfigure}
        \hspace*{\fill}
            \begin{subfigure}{0.45\textwidth}
                \includegraphics[width=0.95\linewidth, clip, trim=0.0cm 5.30cm 0.0cm 0.0cm]{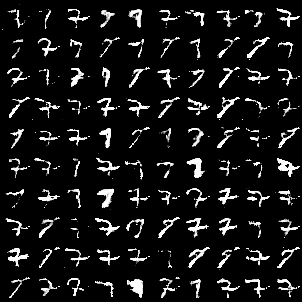}
                \caption{Intended true label `7'.} 
            \end{subfigure}
        \medskip
            \begin{subfigure}{0.45\textwidth}
                \includegraphics[width=0.95\linewidth, clip, trim=0.0cm 5.30cm 0.0cm 0.0cm]{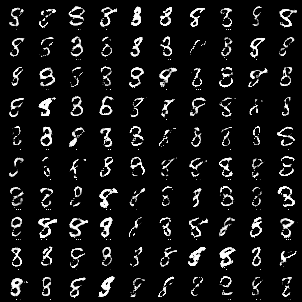}
                \caption{Intended true label `8'.} 
            \end{subfigure}
        \hspace*{\fill}
            \begin{subfigure}{0.45\textwidth}
                \includegraphics[width=0.95\linewidth, clip, trim=0.0cm 5.30cm 0.0cm 0.0cm]{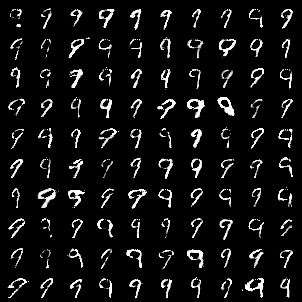}
                \caption{Intended true label `9'.} 
            \end{subfigure}
        \medskip
 \caption{Examples generated 
 	by one adversarially-finetuned GAN
 to perform an untargeted attack on
 \citeauthor{DBLP:conf/icml/WongK18}'s
 (\citeyear{DBLP:conf/icml/WongK18}) classifier,
 which is provably robust to perturbation attacks.}
\end{figure}

\newpage

\vspace{1cm}
\section{Targeted classifiers}
\label{classifier-details}
All targeted classifiers
(other than `simple fully-connected')
are provably robust to adversarial
perturbations in the sense that
there is guaranteed to be
no adversarial input within a distance $\epsilon$
of $p$\% of test inputs under the $l_\infty$ norm.
\vspace{2mm}

\begin{table}[h]
		\setlength{\tabcolsep}{5.5pt} % whitespace between columns
	\renewcommand{\arraystretch}{1.5} % whitespace between rows (multiplicative factor)
		\caption{Descriptions of and references to
		target classifiers used.}
	\begin{tabular}{p{4cm}lllp{5cm}}
		\toprule
		Our Name        & Abbreviation & $\epsilon$ & $p$    & Architecture                                                                    \\
		\midrule
		\cite{DBLP:conf/icml/WongK18}& W\&K         & 0.1 & 94.2 & 2 convolutional layers followed by 2 dense layers                               \\
		MixTrain \citep{DBLP:journals/corr/abs-1811-02625} Model A         & MT-A         & 0.1 & 97.1 & `MNIST\_small': 2 convolutional layers followed by 1 dense layer                \\
		MixTrain \citep{DBLP:journals/corr/abs-1811-02625} Model B         & MT-B         & 0.3 & 60.1 & `MNIST\_small': 2 convolutional layers followed by 1 dense layer                \\
		MixTrain \citep{DBLP:journals/corr/abs-1811-02625} Model C         & MT-C         & 0.1 & 96.4 & `MNIST\_large': 4 convolutional layers followed by 2 dense layers               \\
		MixTrain \citep{DBLP:journals/corr/abs-1811-02625} Model D         & MT-D         & 0.3 & 58.4 & `MNIST\_large': 4 convolutional layers followed by 2 dense layers               \\
		Simple Fully-Connected       & Simple       & N/A & N/A  & Three fully-connected layers of size 256, 128 and 32 with LeakyReLU activations \\
		\bottomrule
	\end{tabular}
\end{table}

\vspace{1cm}

\section{Adversarial Training Experiment}
\label{adv-train-exp-appendix}
The classifier trained during adversarial
training (both the architecture and hyperparameters) 
is the one used in 
\cite{DBLP:journals/corr/MadryMSTV17}, 
and in particular from their associated MNIST Adversarial 
Examples Challenge.

For the experiments with our own model, 
we first pre-train the generator. We then 
continue in `training rounds'. First, we
fine-tune against the classifier
for 5000 gradient steps, using the hyperparameters
from Table \ref{all-hyperparams}, but with an
attack rate of 0.4.
Next, we produce 80,000 attacked training examples
(using an untargeted attack), which are added to the
pool of all examples generated so far.
Then, the classifier is
trained on the entirety of the pool of samples 30 times, 
with a batch size of 128. 
Once a training round is completed we start
again, resetting the GAN to how it was
\textit{before} the adversarial finetuning.

For the experiments with \citeauthor{DBLP:conf/nips/SongSKE18}'s
(\citeyear{DBLP:conf/nips/SongSKE18}) model, 
we run 300 training gradient steps for the
 \citeauthor{DBLP:journals/corr/MadryMSTV17} classifier, 
with a batch size of 64. At each step, the
training data is produced by \citeauthor{DBLP:conf/nips/SongSKE18}'s model.
We use their code and the hyperparameters they provide 
for untargeted attacks in Table 4 of their appendix. 

\newpage

\newpage
\section{MNIST Experiments: Architectures and Hyperparameters}
\label{architectures-hyps}
The WGAN-GP \citep{DBLP:conf/nips/GulrajaniAADC17}
and ACGAN \citep{DBLP:conf/icml/OdenaOS17}
architectures
were the starting points for the design of these neural
networks.
Only a small amount of manual hyperparameter tuning
was performed.

The discriminator network
is
a combination of a conditional WGAN-GP critic, which learns
an approximation of the Wasserstein distance between
the generated and training-set conditional
distributions, and
an auxiliary classifier, which predicts
the likelihood of the possible values of $h(x)$.
We combined these two architectures in an attempt
to strengthen the gradient provided to
the generator, helping to generate data
which are both
realistic and for which the true (i.e., human-judged)
labels match the intended true labels.
The critic is given the true label of the data
$h(x)$ to improve its training,
but the auxiliary classifier must not have
access to this information since its purpose is to
predict it.
We therefore split the discriminator $d$ into three
sub-networks. Network $d_0 \colon X \to \mathbb{R}^i$
effectively preprocesses the input, passing an
intermediate representation to the
critic network
$d_1 \colon \mathbb{R}^i \times Y \to \mathbb{R}$
and the auxiliary classifier network
$d_2 \colon \mathbb{R}^i \to \mathbb{R}^{|Y|}$.
In our experiments,
both $d_1$ and $d_2$ were single
fully-connected layers of the
appropriate dimension.
The loss terms from the WGAN-GP
and ACGAN algorithms
are simply summed.
\vspace{3mm}

\begin{table}[htb!]
	\setlength{\tabcolsep}{6pt} % whitespace between columns
	\begin{center}
		
	\caption{Architecture for generator network, $g$.}
	\begin{tabularx}{\linewidth}{lp{.7cm}lp{1.9cm}p{1.8cm}p{0.9cm}l}
		\toprule
		Layer Type & Kernel & Strides & Feature Maps & Batch Norm. & Dropout & Activation \\
		\midrule
		Fully-Connected & N/A & N/A & \hphantom{0}64 & No & 0\hphantom{.00} & ReLU \\
		Transposed Convolution & $5 \times 5$ & $2 \times 2$ & \hphantom{0}32 & Yes & 0.35 & LeakyReLU  \\
		Transposed Convolution & $5 \times 5$ & $2 \times 2$ & \hphantom{00}8 & Yes & 0.35 & LeakyReLU  \\
		Transposed Convolution & $5 \times 5$ & $2 \times 2$ & \hphantom{00}4 & Yes & 0.35 & LeakyReLU \\
		Fully-Connected & N/A & N/A & 784 & No & 0\hphantom{.00} & Tanh \\
		\bottomrule
	\end{tabularx}
	\medskip\medskip\bigskip
	
	\caption{Architecture for discriminator subnetwork, $d_0$.}
	\begin{tabular}{l*{6}{c}}
		\toprule
		{Layer Type} & {Kernel} & {Strides} & {Feature Maps} & {Batch Norm.} & {Dropout} & {Activation Function} \\
		\midrule
		Convolution & $3 \times 3$ & $2 \times 2$ & \hphantom{00}8 & No & 0.2 & LeakyReLU  \\
		Convolution & $3 \times 3$ & $1 \times 1$ & \hphantom{0}16 & No & 0.2 & LeakyReLU  \\
		Convolution & $3 \times 3$ & $2 \times 2$ & \hphantom{0}32 & No & 0.2 & LeakyReLU  \\
		Convolution & $3 \times 3$ & $1 \times 1$ & \hphantom{0}64 & No & 0.2 & LeakyReLU  \\
		Convolution & $3 \times 3$ & $2 \times 2$ & 128 & No & 0.2 & LeakyReLU  \\
		Convolution & $3 \times 3$ & $1 \times 1$ & 256 & No & 0.2 & LeakyReLU  \\
		\bottomrule
		
	\end{tabular}
	\medskip\medskip\bigskip

	\caption{Hyperparameters for all networks.}
	\begin{tabular}{rl}
		 \toprule
		 {Hyperparameter} & {Value} \\
		 \midrule
		 Attack rate & $\mu = 0.1$ \\
		 Learning rate  &$\alpha= 0.000005$ \\
		 Adam betas & {$\beta_1 = 0.6, \beta_2 = 0.999$} \\
		 Leaky ReLU slope & 0.2 \\
		 Minibatch size & {100} \\
		 Dimensionality of latent space & 128 \\
		 Weight initialisation  & {Normally distributed as described by ~\cite{DBLP:conf/iccv/HeZRS15}} \\
		 Coefficient of gradient penalty loss term & $\lambda =10 $\\
		 
		\bottomrule
		\label{all-hyperparams}
	\end{tabular}
	\medskip
	
	\end{center}
\end{table}

\clearpage

\newpage

\section{Visual Effect of Adversarial Finetuning}

\vspace{0.8cm}

\begin{figure}[htb!]
	\begin{subfigure}{0.45\textwidth}
		\includegraphics[width=\linewidth, clip, trim=0.0cm 0.0cm 0.0cm 4.3cm]{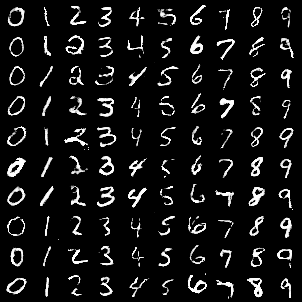}
		\caption{Samples from pretrained generator.} \label{fig:vfta}
	\end{subfigure}\hspace*{\fill}
	\begin{subfigure}{0.45\textwidth}
		\includegraphics[width=\linewidth, clip, trim=0.0cm 0.0cm 0.0cm 4.3cm]{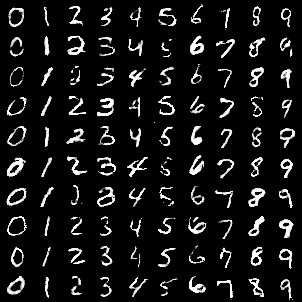}
		\caption{After 5,000 iterations of finetuning.} \label{fig:vftb}
	\end{subfigure}
	
	\medskip
	\begin{subfigure}{0.45\textwidth}
		\includegraphics[width=\linewidth, clip, trim=0.0cm 0.0cm 0.0cm 4.3cm]{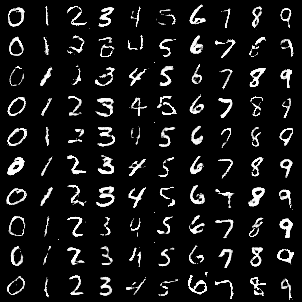}
		\caption{After 10,000 iterations of finetuning.} \label{fig:vftc}
	\end{subfigure}\hspace*{\fill}
	\begin{subfigure}{0.45\textwidth}
		\includegraphics[width=\linewidth, clip, trim=0.0cm 0.0cm 0.0cm 4.3cm]{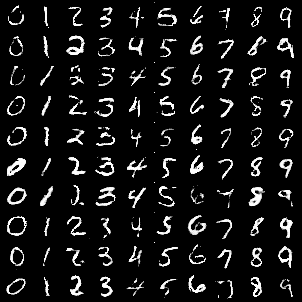}
		\caption{After 20,000 iterations of finetuning.} \label{fig:vftd}
	\end{subfigure}
	
	\medskip
	\begin{subfigure}{0.45\textwidth}
		\includegraphics[width=\linewidth, clip, trim=0.0cm 0.0cm 0.0cm 4.3cm]{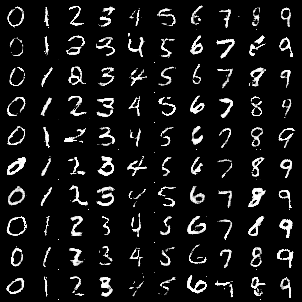}
		\caption{After 30,000 iterations of finetuning.} \label{fig:vfte}
	\end{subfigure}\hspace*{\fill}
	\begin{subfigure}{0.45\textwidth}
		\includegraphics[width=\linewidth, clip, trim=0.0cm 0.0cm 0.0cm 4.3cm]{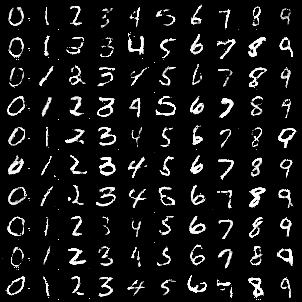}
		\caption{After 45,000 iterations of finetuning.
				We ended finetuning
			at this stage.} \label{fig:vftf}
	\end{subfigure}
	
	\caption{A sequence of images tracking the output of the
		generator network for one fixed random sample in
		latent space as adversarial finetuning takes place.
		Five samples are given for each intended true label.
		The finetuning is an untargeted attack against
		\citeauthor{DBLP:conf/icml/WongK18}'s
		(\citeyear{DBLP:conf/icml/WongK18})
		provably-robust network.} \label{fig:visual-finetuning}
\end{figure}

\clearpage

\section{Transferability of Adversarial Examples}

Perturbation-based
adversarial examples typically somewhat generalise
between models
\citep{DBLP:journals/corr/SzegedyZSBEGF13,
	DBLP:conf/iclr/LiuCLS17}.
That is, inputs crafted using white-box
access to fool one model often fool a
different model.
This means that black-box attacks are possible, if
the attacker has a different trained model for the same task.
To evaluate whether our method could be used
in the same way, we generated about 20,000 untargeted
unrestricted adversarial inputs for each target
classifier, and measured the misclassification
rates on this set for the other models.
The high variance
of the results, shown in Table~\ref{transfer-table},
suggests that successful transfer may depend
more on the networks in question than
on our generation algorithm.

\begin{table}[h]
	\centering
	\caption{The percentage of adversarial
		examples targeting each classifier which are also adversarial for the others.
	See Appendix~\ref{classifier-details}
	for descriptions of the classifiers.}
	\label{transfer-table}
	\setlength{\tabcolsep}{2.5pt} % whitespace between columns
	\renewcommand{\arraystretch}{1.1} % whitespace between rows (multiplicative factor)
	\vspace{1mm}
	\begin{tabular}{p{0.6cm}lcccccc}
		&        &      &        & To     &        &        &        \\
		\toprule
		&        & W\&K   & MT-A & MT-B & MT-C & MT-D & Simple \\
		\midrule
		& W\&K     &      & 20.2   & 18.4   & \hphantom{0}9.0      & 60.7   & 16.8   \\
		& MT-A & 19.5 &        & 14.1   & 13.3   & 55.2   & \hphantom{0}4.7    \\
		\parbox[t]{0cm}{\multirow{8}{*}{\rotatebox[origin=c]{90}{\hspace{4cm}From}}} & MT-B & \hphantom{0}5.2  & \hphantom{0}4.8    &        & \hphantom{0}1.6    & 57.8   & \hphantom{0}2.6    \\
		& MT-C & 25.8 & 47.6   & 13.9   &        & 67.8   & 12.1   \\
		& MT-D & \hphantom{0}5.9  & \hphantom{0}7.3    & \hphantom{0}9.4    & \hphantom{0}4.3    &        & \hphantom{0}1.7    \\
		& Simple & \hphantom{0}2.7  & \hphantom{0}2.6    & \hphantom{0}2.6    & \hphantom{0}1.3    & 48.0     &    \\
		\bottomrule
	\end{tabular}
\end{table}

\vspace{1cm}

\section{Safeguarding MTurk Data Quality}
\label{mturk-app}
The evaluation of our method relies entirely on the quality of the data provided by the MTurk workers. We
therefore took a number of measures to ensure that participants understood the instructions and completed the
tasks diligently:
\begin{itemize}
	\item Only workers with good track records were permitted to participate.
	\item The instructions specified that particular answers
	should be given to specified questions to prove that
	the instructions had been read carefully. Approximately 10\% of work was rejected for failing this check.
	\item For the image labelling tasks, some images with
	known labels were included to check that the right
	labels were being given. Reassuringly, almost no
	work was rejected for failing this check.
	\item For the identification of the generated images, a bonus nearly doubling the pay per image was given
	for each correctly-identified image, providing an
	extra incentive to try hard.
	\item To provide a disincentive to high-speed random
	clicking, a minimum time spent answering each
	question was enforced.
	\item If more than 1\% of questions were left unanswered, we interpreted
	this as a sign of carelessness and did not use any
	of the data from that task.
\end{itemize}

\clearpage

\section{Results for Non-Robust Target Network}
\label{nonrobust-results}
These results are targeting a simple convolutional
neural network 
with LeakyReLU activations and three hidden layers of size
256, 128 and 32, trained until convergence.
\vspace{1cm}
\begin{figure}[H]
    \begin{minipage}[t]{0.480\linewidth}\centering
    \hspace{1cm} Target label 

    \begin{tabular}{rr*{11}{L}}
    
 & & $0$ & $1$ & $2$ & $3$ & $4$ & $5$ & $6$ & $7$ & $8$ & $9$ & None \\
\parbox[t]{-9cm}{\multirow{11}{*}{\rotatebox[origin=c]{90}{Intended true label}}} & 0 \hspace{0.8mm} & & 93  & 96  & 96  & 99  & 93  & 95  & 97  & 94  & 97  & 96  \\
& 1 \hspace{0.8mm} & XX  & & 92  & 100  & 92  & 97  & 96  & 88  & 96  & 96  & 93  \\
& 2 \hspace{0.8mm} & 73  & 86  & & 82  & 80  & 87  & 92  & 84  & 87  & 75  & 81  \\
& 3 \hspace{0.8mm} & 88  & 83  & 87  & & 81  & 88  & 81  & 89  & 96  & 90  & 92  \\
& 4 \hspace{0.8mm} & 84  & 53  & 79  & 69  & & 78  & 90  & 90  & 81  & 87  & 85  \\
& 5 \hspace{0.8mm} & 84  & 89  & 77  & 89  & 88  & & 79  & 94  & 88  & 88  & 83  \\
& 6 \hspace{0.8mm} & 96  & 83  & 92  & 95  & 93  & 95  & & 93  & 100  & 96  & 94  \\
& 7 \hspace{0.8mm} & 93  & 59  & 89  & 95  & 85  & 94  & 80  & & 99  & 94  & 92  \\
& 8 \hspace{0.8mm} & 96  & 86  & 97  & 93  & 98  & 93  & 90  & 92  & & 92  & 91  \\
& 9 \hspace{0.8mm} & 93  & 76  & 96  & 97  & 97  & 91  & 89  & 93  & 89  & & 98  \\
 &  &  &  &  &  &  &  &  &  &  & \\[-3mm]
& Mean \hspace{0.6mm} & 88 & 79 & 89 & 91 & 90 & 91 & 88 & 91 & 92 & 91 & 90\\\end{tabular}
 \caption{The success rates of the adversarial attacks by finetuned GANs. More precisely, of generated images for which the computed label output by the classifier matches the target label, the percentage which are truly adversarial (in the sense that the true label of the image matches the intended true label passed to the generator network) is reported.} 
\label{labelme-mnist-finetuned-normal}
\end{minipage}
\hfill{}
    \begin{minipage}[t]{0.480\linewidth}\centering
    \hspace{1cm} Target label 

    \begin{tabular}{rr*{11}{L}}
    
 & & $0$ & $1$ & $2$ & $3$ & $4$ & $5$ & $6$ & $7$ & $8$ & $9$ & None \\
\parbox[t]{-9cm}{\multirow{11}{*}{\rotatebox[origin=c]{90}{Intended true label}}} & 0 \hspace{0.8mm} & & XX  & 76  & 44  & 65  & 70  & 77  & 89  & 56  & 75  & 76  \\
& 1 \hspace{0.8mm} & XX  & & 82  & 89  & 95  & 99  & 93  & 76  & 98  & 98  & 98  \\
& 2 \hspace{0.8mm} & 35  & 45  & & 94  & 55  & 40  & 55  & 75  & 75  & 48  & 68  \\
& 3 \hspace{0.8mm} & 64  & 66  & 66  & & 46  & 71  & 33  & 73  & 81  & 84  & 80  \\
& 4 \hspace{0.8mm} & 66  & 41  & 66  & 68  & & 56  & 66  & 61  & 59  & 85  & 77  \\
& 5 \hspace{0.8mm} & 64  & 68  & 80  & 76  & 71  & & 69  & 63  & 65  & 82  & 73  \\
& 6 \hspace{0.8mm} & 82  & 63  & 66  & 43  & 91  & 79  & & 78  & 81  & 60  & 85  \\
& 7 \hspace{0.8mm} & 73  & 47  & 82  & 80  & 72  & 92  & 100  & & 92  & 86  & 77  \\
& 8 \hspace{0.8mm} & 87  & 69  & 81  & 77  & 79  & 65  & 88  & 62  & & 81  & 85  \\
& 9 \hspace{0.8mm} & 90  & 45  & 79  & 75  & 76  & 96  & XX  & 82  & 74  & & 84  \\
 &  &  &  &  &  &  &  &  &  &  & \\[-3mm]
& Mean \hspace{0.6mm} & 70 & 56 & 75 & 72 & 72 & 74 & 73 & 73 & 76 & 78 & 80\\\end{tabular}
 \caption{The success rates of the adversarial attacks by a pretrained but not finetuned GAN. More precisely, of generated images for which the computed label output by the classifier matches the target label, the percentage which are truly adversarial (in the sense that the true label of the image matches the intended true label passed to the generator network) is reported.} 
\label{labelme-mnist-pretrained-normal}
\end{minipage}
\medskip\medskip

    \begin{minipage}[t]{0.480\linewidth}\centering
    \hspace{1cm} Target label 

    \begin{tabular}{rr*{11}{O}}
    
 & & $0$ & $1$ & $2$ & $3$ & $4$ & $5$ & $6$ & $7$ & $8$ & $9$ & None \\
\parbox[t]{-9cm}{\multirow{11}{*}{\rotatebox[origin=c]{90}{Intended true label}}} & 0 \hspace{0.8mm} & & 45  & 48  & 46  & 43  & 38  & 61  & 51  & 44  & 59  & 53  \\
& 1 \hspace{0.8mm} & XX  & & 64  & 74  & 72  & 62  & 78  & 83  & 73  & 75  & 79  \\
& 2 \hspace{0.8mm} & 52  & 43  & & 53  & 41  & 52  & 35  & 48  & 54  & 40  & 55  \\
& 3 \hspace{0.8mm} & 64  & 41  & 62  & & 43  & 60  & 29  & 51  & 56  & 58  & 50  \\
& 4 \hspace{0.8mm} & 59  & 45  & 49  & 36  & & 45  & 53  & 55  & 49  & 69  & 65  \\
& 5 \hspace{0.8mm} & 48  & 46  & 44  & 63  & 62  & & 60  & 49  & 57  & 61  & 50  \\
& 6 \hspace{0.8mm} & 76  & 48  & 44  & 46  & 54  & 54  & & 38  & 62  & 54  & 58  \\
& 7 \hspace{0.8mm} & 51  & 32  & 60  & 59  & 54  & 54  & 46  & & 61  & 65  & 65  \\
& 8 \hspace{0.8mm} & 62  & 53  & 62  & 57  & 56  & 56  & 54  & 50  & & 67  & 57  \\
& 9 \hspace{0.8mm} & 47  & 42  & 51  & 60  & 72  & 69  & 54  & 66  & 71  & & 67  \\
 &  &  &  &  &  &  &  &  &  &  & \\[-3mm]
& Mean \hspace{0.6mm} & 57 & 44 & 54 & 55 & 55 & 54 & 52 & 55 & 59 & 61 & 60\\\end{tabular}
 \caption{Measures of how realistic the adversarial images generated by finetuned GANs are. More precisely, the proportion of generated inputs for which the classified label matches the target label which were not identified as being generated when placed amongst nine images from the training dataset. If the generated images were completely realistic, the expected result would be 90.} 
\label{oddoneout-mnist-finetuned-normal}
\end{minipage}
\hfill{}
    \begin{minipage}[t]{0.480\linewidth}\centering
    \hspace{1cm} Target label 

    \begin{tabular}{rr*{11}{O}}
    
 & & $0$ & $1$ & $2$ & $3$ & $4$ & $5$ & $6$ & $7$ & $8$ & $9$ & None \\
\parbox[t]{-9cm}{\multirow{11}{*}{\rotatebox[origin=c]{90}{Intended true label}}} & 0 \hspace{0.8mm} & & XX  & 59  & 56  & 61  & 65  & 63  & 55  & 51  & 54  & 63  \\
& 1 \hspace{0.8mm} & XX  & & 56  & 62  & 76  & 76  & 70  & 72  & 80  & 77  & 74  \\
& 2 \hspace{0.8mm} & 53  & 55  & & 66  & 54  & 48  & 50  & 62  & 64  & 52  & 61  \\
& 3 \hspace{0.8mm} & 68  & 54  & 62  & & 48  & 64  & 52  & 64  & 71  & 65  & 71  \\
& 4 \hspace{0.8mm} & 57  & 54  & 55  & 57  & & 63  & 52  & 60  & 47  & 66  & 73  \\
& 5 \hspace{0.8mm} & 69  & 64  & 58  & 62  & 57  & & 73  & 53  & 60  & 62  & 63  \\
& 6 \hspace{0.8mm} & 63  & 64  & 58  & 59  & 67  & 71  & & 62  & 63  & 63  & 67  \\
& 7 \hspace{0.8mm} & 54  & 63  & 71  & 61  & 71  & 62  & 52  & & 67  & 74  & 77  \\
& 8 \hspace{0.8mm} & 65  & 60  & 60  & 71  & 57  & 60  & 81  & 55  & & 69  & 67  \\
& 9 \hspace{0.8mm} & 64  & 51  & 65  & 64  & 81  & 71  & XX  & 76  & 68  & & 68  \\
 &  &  &  &  &  &  &  &  &  &  & \\[-3mm]
& Mean \hspace{0.6mm} & 62 & 58 & 60 & 62 & 64 & 64 & 62 & 62 & 63 & 65 & 68\\\end{tabular}
 \caption{Measures of how realistic the adversarial images generated by a pretrained but not finetuned GAN are. More precisely, the proportion of generated inputs for which the classified label matches the target label which were not identified as being generated when placed amongst nine images from the training dataset. If the generated images were completely realistic, the expected result would be 90.} 
\label{oddoneout-mnist-pretrained-normal}
\end{minipage}
\end{figure}

\clearpage

\section{Side-by-Side Image Comparison Results}
\label{abtest-results}
Each figure shows the number of human
judgements out of 100 which correctly
identified the unrestricted adversarial input
in a side-by-side comparison with an image
drawn from the dataset.
If the generated images were completely realistic,
the expected result would be 50.
\vspace{1cm}

\begin{figure}[H]
    \begin{minipage}[t]{0.480\linewidth}\centering
    \hspace{1cm} Target label 

    \begin{tabular}{rr*{11}{A}}
    
 & & $0$ & $1$ & $2$ & $3$ & $4$ & $5$ & $6$ & $7$ & $8$ & $9$ & None \\
\parbox[t]{-9cm}{\multirow{11}{*}{\rotatebox[origin=c]{90}{Intended true label}}} & 0 \hspace{0.8mm} & & 23  & 29  & 24  & 30  & 25  & 20  & 22  & 17  & 22  & 28  \\
& 1 \hspace{0.8mm} & XX  & & 14  & 19  & 16  & 26  & 38  & 25  & 21  & 21  & 24  \\
& 2 \hspace{0.8mm} & 24  & 25  & & 18  & 25  & 21  & 17  & 26  & 17  & 19  & 29  \\
& 3 \hspace{0.8mm} & 22  & 24  & 17  & & 31  & 28  & 25  & 24  & 24  & 30  & 20  \\
& 4 \hspace{0.8mm} & 25  & 25  & 28  & 20  & & 21  & 28  & 22  & 17  & 23  & 19  \\
& 5 \hspace{0.8mm} & 23  & 16  & 24  & 27  & 29  & & 23  & 19  & 27  & 21  & 21  \\
& 6 \hspace{0.8mm} & 19  & 21  & 25  & 21  & 19  & 25  & & 20  & 28  & 18  & 23  \\
& 7 \hspace{0.8mm} & 23  & 27  & 22  & 26  & 17  & 24  & 25  & & 29  & 16  & 21  \\
& 8 \hspace{0.8mm} & 25  & 25  & 21  & 21  & 24  & 23  & 24  & 25  & & 28  & 23  \\
& 9 \hspace{0.8mm} & 18  & 21  & 22  & 27  & 27  & 24  & 23  & 28  & 23  & & 28  \\
 &  &  &  &  &  &  &  &  &  &  & \\[-3mm]
& Mean \hspace{0.6mm} & 22 & 23 & 22 & 23 & 24 & 24 & 25 & 23 & 23 & 22 & 24\\\end{tabular}
 \caption{Results against \cite{DBLP:conf/icml/WongK18} generated by adversarially finetuned GANs.} 
\label{abtest-mnist-finetuned-robust}
\end{minipage}
\hfill{}
    \begin{minipage}[t]{0.480\linewidth}\centering
    \hspace{1cm} Target label 

    \begin{tabular}{rr*{11}{A}}
    
 & & $0$ & $1$ & $2$ & $3$ & $4$ & $5$ & $6$ & $7$ & $8$ & $9$ & None \\
\parbox[t]{-9cm}{\multirow{11}{*}{\rotatebox[origin=c]{90}{Intended true label}}} & 0 \hspace{0.8mm} & & 43  & 37  & 42  & 42  & 41  & 36  & 43  & 38  & 45  & 36  \\
& 1 \hspace{0.8mm} & XX  & & 29  & 40  & 38  & 40  & 39  & 40  & 43  & 37  & 34  \\
& 2 \hspace{0.8mm} & 45  & 30  & & 38  & 36  & 35  & 39  & 44  & 41  & 31  & 40  \\
& 3 \hspace{0.8mm} & 35  & 35  & 38  & & 36  & 45  & 34  & 42  & 44  & 38  & 36  \\
& 4 \hspace{0.8mm} & 30  & 42  & 42  & 40  & & 36  & 35  & 47  & 42  & 38  & 42  \\
& 5 \hspace{0.8mm} & 35  & 43  & 40  & 35  & 34  & & 34  & 42  & 36  & 37  & 44  \\
& 6 \hspace{0.8mm} & 44  & 46  & 46  & 42  & 37  & 42  & & 45  & 41  & 38  & 36  \\
& 7 \hspace{0.8mm} & 32  & 38  & 38  & 42  & 42  & 43  & 33  & & 35  & 46  & 45  \\
& 8 \hspace{0.8mm} & 41  & 41  & 44  & 46  & 37  & 37  & 36  & 43  & & 42  & 44  \\
& 9 \hspace{0.8mm} & 45  & 41  & 35  & 47  & 44  & 40  & 40  & 52  & 36  & & 49  \\
 &  &  &  &  &  &  &  &  &  &  & \\[-3mm]
& Mean \hspace{0.6mm} & 38 & 40 & 39 & 41 & 38 & 40 & 36 & 44 & 40 & 39 & 41\\\end{tabular}
 \caption{Results against \cite{DBLP:conf/icml/WongK18} generated by a pretrained but not finetuned GAN.} 
\label{abtest-mnist-pretrained-robust}
\end{minipage}
\medskip\medskip

    \begin{minipage}[t]{0.480\linewidth}\centering
    \hspace{1cm} Target label 

    \begin{tabular}{rr*{11}{A}}
    
 & & $0$ & $1$ & $2$ & $3$ & $4$ & $5$ & $6$ & $7$ & $8$ & $9$ & None \\
\parbox[t]{-9cm}{\multirow{11}{*}{\rotatebox[origin=c]{90}{Intended true label}}} & 0 \hspace{0.8mm} & & 23  & 32  & 23  & 22  & 22  & 23  & 19  & 24  & 27  & 23  \\
& 1 \hspace{0.8mm} & XX  & & 31  & 32  & 31  & 28  & 30  & 36  & 41  & 29  & 28  \\
& 2 \hspace{0.8mm} & 24  & 28  & & 31  & 22  & 24  & 30  & 26  & 31  & 20  & 28  \\
& 3 \hspace{0.8mm} & 30  & 26  & 30  & & 20  & 20  & 25  & 19  & 21  & 25  & 23  \\
& 4 \hspace{0.8mm} & 27  & 28  & 25  & 20  & & 28  & 27  & 25  & 26  & 27  & 30  \\
& 5 \hspace{0.8mm} & 29  & 25  & 25  & 25  & 24  & & 23  & 25  & 22  & 22  & 28  \\
& 6 \hspace{0.8mm} & 30  & 22  & 27  & 17  & 30  & 28  & & 20  & 32  & 23  & 29  \\
& 7 \hspace{0.8mm} & 18  & 24  & 30  & 32  & 25  & 24  & 22  & & 34  & 26  & 34  \\
& 8 \hspace{0.8mm} & 28  & 29  & 19  & 23  & 22  & 28  & 27  & 26  & & 26  & 37  \\
& 9 \hspace{0.8mm} & 29  & 21  & 20  & 31  & 28  & 33  & 21  & 44  & 30  & & 38  \\
 &  &  &  &  &  &  &  &  &  &  & \\[-3mm]
& Mean \hspace{0.6mm} & 27 & 25 & 27 & 26 & 25 & 26 & 25 & 27 & 29 & 25 & 30\\\end{tabular}
 \caption{Results against an ordinary neural network generated by adversarially finetuned GANs.} 
\label{abtest-mnist-finetuned-normal}
\end{minipage}
\hfill{}
    \begin{minipage}[t]{0.480\linewidth}\centering
    \hspace{1cm} Target label 

    \begin{tabular}{rr*{11}{A}}
    
 & & $0$ & $1$ & $2$ & $3$ & $4$ & $5$ & $6$ & $7$ & $8$ & $9$ & None \\
\parbox[t]{-9cm}{\multirow{11}{*}{\rotatebox[origin=c]{90}{Intended true label}}} & 0 \hspace{0.8mm} & & XX  & 25  & 28  & 31  & 26  & 34  & 16  & 20  & 21  & 27  \\
& 1 \hspace{0.8mm} & XX  & & 36  & 26  & 31  & 27  & 25  & 24  & 38  & 22  & 25  \\
& 2 \hspace{0.8mm} & 24  & 24  & & 29  & 28  & 26  & 21  & 28  & 25  & 22  & 27  \\
& 3 \hspace{0.8mm} & 23  & 27  & 26  & & 26  & 29  & 23  & 22  & 29  & 31  & 31  \\
& 4 \hspace{0.8mm} & 27  & 18  & 23  & 31  & & 28  & 24  & 29  & 32  & 37  & 33  \\
& 5 \hspace{0.8mm} & 26  & 24  & 30  & 24  & 29  & & 28  & 23  & 32  & 30  & 30  \\
& 6 \hspace{0.8mm} & 37  & 23  & 23  & 21  & 28  & 30  & & 26  & 29  & 25  & 26  \\
& 7 \hspace{0.8mm} & 23  & 33  & 22  & 29  & 26  & 25  & 24  & & 27  & 28  & 32  \\
& 8 \hspace{0.8mm} & 31  & 20  & 21  & 26  & 29  & 31  & 31  & 26  & & 33  & 30  \\
& 9 \hspace{0.8mm} & 27  & 26  & 26  & 22  & 32  & 26  & XX  & 31  & 26  & & 30  \\
 &  &  &  &  &  &  &  &  &  &  & \\[-3mm]
& Mean \hspace{0.6mm} & 27 & 24 & 26 & 26 & 29 & 28 & 26 & 25 & 29 & 28 & 29\\\end{tabular}
 \caption{Results against an ordinary neural network generated by a pretrained but not finetuned GAN.} 
\label{abtest-mnist-pretrained-normal}
\end{minipage}
\end{figure}

\newpage

\section{Results for Pretrained Baseline}
\label{pretrained-results}
These results are for data
generated by a GAN which has been pretrained
but not adversarially finetuned, targeting
\cite{DBLP:conf/icml/WongK18}
provably-robust network.
\vspace{1cm}

\begin{figure}[h]
    \begin{minipage}[t]{0.480\linewidth}\centering
    \hspace{1cm} Target label 

    \begin{tabular}{rr*{11}{L}}
    
 & & $0$ & $1$ & $2$ & $3$ & $4$ & $5$ & $6$ & $7$ & $8$ & $9$ & None \\
\parbox[t]{-9cm}{\multirow{11}{*}{\rotatebox[origin=c]{90}{Intended true label}}} & 0 \hspace{0.8mm} & & 73  & 91  & 62  & 62  & 91  & 77  & 88  & 50  & 59  & 75  \\
& 1 \hspace{0.8mm} & XX  & & 49  & 51  & 80  & 06  & 31  & 47  & 61  & 77  & 57  \\
& 2 \hspace{0.8mm} & 13  & 62  & & 53  & 32  & 30  & 30  & 52  & 50  & 19  & 60  \\
& 3 \hspace{0.8mm} & 29  & 69  & 60  & & 26  & 60  & 12  & 79  & 22  & 28  & 71  \\
& 4 \hspace{0.8mm} & 42  & 55  & 66  & 43  & & 70  & 80  & 70  & 48  & 73  & 65  \\
& 5 \hspace{0.8mm} & 18  & 46  & 55  & 61  & 54  & & 29  & 59  & 31  & 51  & 60  \\
& 6 \hspace{0.8mm} & 50  & 60  & 80  & 87  & 74  & 80  & & 74  & 54  & 59  & 66  \\
& 7 \hspace{0.8mm} & 22  & 43  & 82  & 63  & 31  & 64  & 00  & & 62  & 38  & 40  \\
& 8 \hspace{0.8mm} & 70  & 68  & 80  & 75  & 75  & 91  & 63  & 50  & & 80  & 79  \\
& 9 \hspace{0.8mm} & 55  & 66  & 88  & 74  & 88  & 92  & 66  & 87  & 69  & & 88  \\
 &  &  &  &  &  &  &  &  &  &  & \\[-3mm]
& Mean \hspace{0.6mm} & 37 & 60 & 72 & 63 & 58 & 65 & 43 & 67 & 50 & 54 & 66\\\end{tabular}
 \caption{The success rates of the adversarial attacks by a pretrained but not finetuned GAN. More precisely, of generated images for which the computed label output by the classifier matches the target label, the percentage which are truly adversarial (in the sense that the true label of the image matches the intended true label passed to the generator network) is reported.} 
\label{labelme-mnist-pretrained-robust}
\end{minipage}
\hfill{}
    \begin{minipage}[t]{0.480\linewidth}\centering
    \hspace{1cm} Target label 

    \begin{tabular}{rr*{11}{O}}
    
 & & $0$ & $1$ & $2$ & $3$ & $4$ & $5$ & $6$ & $7$ & $8$ & $9$ & None \\
\parbox[t]{-9cm}{\multirow{11}{*}{\rotatebox[origin=c]{90}{Intended true label}}} & 0 \hspace{0.8mm} & & 48  & 63  & 53  & 53  & 62  & 56  & 46  & 40  & 37  & 54  \\
& 1 \hspace{0.8mm} & XX  & & 34  & 52  & 50  & 37  & 27  & 52  & 47  & 46  & 42  \\
& 2 \hspace{0.8mm} & 42  & 39  & & 51  & 39  & 30  & 36  & 54  & 49  & 40  & 43  \\
& 3 \hspace{0.8mm} & 41  & 47  & 55  & & 42  & 55  & 47  & 53  & 41  & 47  & 51  \\
& 4 \hspace{0.8mm} & 49  & 52  & 41  & 46  & & 45  & 50  & 51  & 45  & 66  & 57  \\
& 5 \hspace{0.8mm} & 38  & 50  & 43  & 56  & 47  & & 54  & 44  & 51  & 48  & 51  \\
& 6 \hspace{0.8mm} & 49  & 52  & 47  & 50  & 46  & 55  & & 38  & 62  & 46  & 48  \\
& 7 \hspace{0.8mm} & 39  & 57  & 59  & 36  & 49  & 45  & 32  & & 41  & 57  & 59  \\
& 8 \hspace{0.8mm} & 52  & 50  & 59  & 52  & 56  & 57  & 43  & 39  & & 51  & 59  \\
& 9 \hspace{0.8mm} & 51  & 51  & 66  & 53  & 74  & 61  & 53  & 73  & 48  & & 68  \\
 &  &  &  &  &  &  &  &  &  &  & \\[-3mm]
& Mean \hspace{0.6mm} & 45 & 50 & 52 & 50 & 51 & 50 & 44 & 50 & 47 & 49 & 53\\\end{tabular}
 \caption{Measures of how realistic the adversarial images generated by a pretrained but not finetuned GAN are. More precisely, the proportion of generated inputs for which the classified label matches the target label which were not identified as being generated when placed amongst nine images from the training dataset. If the generated images were completely realistic, the expected result would be 90.} 
\label{oddoneout-mnist-pretrained-robust}
\end{minipage}
\end{figure}

\end{document}